%% file: main.tex
\documentclass[11pt]{article}

\usepackage[final]{acl}
\usepackage{times}
\usepackage{latexsym}
\usepackage[T1]{fontenc}
\usepackage[utf8]{inputenc}
\usepackage{microtype}
\usepackage{inconsolata}
\usepackage{graphicx}
\usepackage{amsmath}
\usepackage{amssymb}
\usepackage{amsfonts}
\usepackage{mathtools}
\usepackage{booktabs}
\usepackage{nicefrac}
\usepackage{enumitem}
\usepackage{multirow}
\usepackage[table]{xcolor}
\usepackage[normalem]{ulem}
\usepackage{url}
\usepackage{array}

\definecolor{ourblue}{HTML}{DDEBFF}
\definecolor{regstrong}{HTML}{E6F0FF}
\definecolor{regbypass}{HTML}{FFF0D9}
\definecolor{reginsens}{HTML}{EEEEEE}
\newcommand{\mvh}{\textsc{MVH}}

\newcommand{\kzero}{\textsc{K=0}}
\newcommand{\nolatent}{\textsc{NO\_LATENT}}

\title{Beyond Visual Memory: \\Mechanistic Diagnostics of Latent Visual Reasoning}
\author{
{\large\bfseries Jiawei Guo\textsuperscript{1,2,3},
Yu Chen\textsuperscript{3},
Xiang Wang\textsuperscript{3},
Shuai Li\textsuperscript{3},
Xinpei Zhao\textsuperscript{3}} \\
{\large\bfseries Huaxing Liu\textsuperscript{3},
Shuai Dong\textsuperscript{4},
Feifei Zhai\textsuperscript{1}\footnotemark[2],
Yu Zhou\textsuperscript{1,5}} \\
{\normalsize\mdseries $^1$ State Key Laboratory of Multimodal Artificial Intelligence Systems,} \\
{\normalsize\mdseries Institute of Automation, CAS, Beijing, China} \\
{\normalsize\mdseries $^2$ School of Artificial Intelligence, University of Chinese Academy of Sciences, Beijing, China} \\
{\normalsize\mdseries $^3$ Amap, Alibaba Group \quad $^4$ Shanghai Innovation Institute} \\
{\normalsize\mdseries $^5$ Fanyu AI Laboratory, Zhongke Fanyu Technology Co., Ltd, Beijing, China} \\
{\normalsize\mdseries \{guojiawei2024, feifei.zhai\}@ia.ac.cn,
yzhou@nlpr.ia.ac.cn}
}

\begin{document}
\maketitle

\renewcommand{\thefootnote}{\fnsymbol{footnote}}
\footnotetext[2]{\ \ Corresponding author.}

\input{0_abstract}
\input{1_introduction}
\input{2_related}
\input{3_method}

\input{4_results}
\input{5_analysis}
\input{6_conclusion}

\section*{Limitations}
\label{sec:limitations}

Our analysis focuses on latent visual reasoning methods that use a
full-image VLM interface and insert latent slots into the decoding
sequence. Architectures that enforce a stronger visual bottleneck, restrict
image access during answer generation, or use explicit external visual
workspaces may show different balances between slot contents, boundary
markers, and format. We also study a controlled set of method-stage
settings built around Qwen-family backbones and perception-heavy
benchmarks. Broader model families, video inputs, embodied tasks, or
long-horizon planning may introduce additional uses of latent states that
are not covered here. Finally, our diagnostics are designed to test whether
slot contents act as recoverable visual memory; they do not rule out other
roles for latent slots, such as providing computation, timing, formatting,
or routing capacity during decoding.

\bibliography{custom}

\clearpage
\appendix
\setcounter{table}{0}
\renewcommand{\thetable}{A\arabic{table}}
\setcounter{figure}{0}
\renewcommand{\thefigure}{A\arabic{figure}}

\section{\mvh{}: Full Method Details}
\label{sec:method-appendix}

This appendix gives the equations, training-time assembly, and hyperparameters elided in Section~\ref{sec:mvh-probe}.

\subsection{Problem setup}
\label{app:method-setup}
At each reasoning step $t$, the frozen vision encoder processes a
helper image $x_t^{\text{hlp}}$ (a crop, an annotated view, or a
simulated next state drawn from the training data) and produces a set
of post-merger visual tokens
\[
V_t = \{v_{t,1},\ldots,v_{t,M_t}\},\qquad v_{t,j}\in\mathbb{R}^{d_h},
\]
arranged on an $H_t\times W_t$ spatial grid with $H_t\cdot W_t = M_t$.
From the last window-attention block of the encoder we additionally
extract per-head self-attention maps
$\{A_t^{(h)}\}_{h=1}^{H}$, where
$A_t^{(h)}(i,j)$ is the weight from token $i$ to token $j$ in head
$h$. Both $V_t$ and $\{A_t^{(h)}\}$ are fixed once the helper image is
fixed. We further denote by $h_t\in\mathbb{R}^{d_h}$ the student's
last-layer hidden state at the text token immediately preceding the latent tokens at step $t$; this is the only quantity that changes
step-to-step.

\subsection{Teacher targets: design principle}
\label{app:method-design}
At each step, \mvh{} constructs a structured teacher hypothesis set
\[
T_t = [G_t;\, I_t],\qquad |T_t| = K_g + K_i = N,
\]
organized along two complementary axes. The coverage block $G_t$ is
question-independent and preserves global scene structure
(\S\ref{app:coverage}); the informative block $I_t$ is
question-dependent and rewrites high-information-density anchor
tokens with question-relevant evidence
(\S\ref{app:informative}). The assembly and injection procedures are
given in \S\ref{app:teacher-sequence}--\S\ref{app:norm-align}, and
the training objectives that align student slots to $T_t$ in
\S\ref{app:stage1}--\S\ref{app:stage2}.

\subsection{Coverage states}
\label{app:coverage}
The global mean over visual tokens is
\begin{equation}
g_0 = \frac{1}{M_t}\sum_{j=1}^{M_t} v_{t,j},
\label{eq:g0}
\end{equation}
and $g_1,\ldots,g_9$ are the means of the tokens falling in each cell
of a fixed $3\times 3$ partition of the $H_t\times W_t$ grid. When a
side is not divisible by three, the remainder is assigned to the last
row/column; empty cells fall back to $g_0$. We take
\[
G_t = \{g_0, g_1, \ldots, g_9\},\qquad |G_t| = K_g = 10.
\]
The $3\times 3$ partition is the coarsest grid that can still
express layouts such as a top-right quadrant or a middle-row band,
without the combinatorial blow-up of denser partitions; the global
token $g_0$ provides a grid-agnostic summary.

\subsection{Informative hypotheses: selection and rewriting}
\label{app:informative}
\paragraph{Base informativeness.}
The encoder-side informativeness of token $j$ is the per-head average
of incoming attention,
\begin{equation}
r_j^{\text{base}} = \frac{1}{HM_t}\sum_{h=1}^{H}\sum_{i=1}^{M_t}
A_t^{(h)}(i,j),
\label{eq:r-base}
\end{equation}
followed by an $\ell_1$ renormalization. Tokens with high
$r_j^{\text{base}}$ act as information hubs that absorb and
redistribute compressed global information during encoding.

\paragraph{Query relevance.}
We project the current step's text context $h_t$ against the visual
tokens with a scaled dot product and normalize,
\begin{align}
q_{t,j} &= (v_{t,j})^{\!\top} h_t \,/\, \sqrt{d_h}, \label{eq:q}\\
u_{t,j} &= \frac{\exp(q_{t,j})}{\sum_k \exp(q_{t,k})}. \label{eq:u}
\end{align}
Because $V_t$ and $h_t$ already share the decoder's hidden
dimension, no extra projection matrix is required.

\paragraph{Fused score and top-$K_i$ selection.}
The two signals are combined multiplicatively with equal weight and
renormalized,
\begin{equation}
p_{t,j}
= \frac{r_j^{\text{base}}\cdot u_{t,j}}
       {\sum_{k} r_k^{\text{base}}\cdot u_{t,k}},
\label{eq:fused-score-app}
\end{equation}
and we take $I_t^{\text{idx}} = \mathrm{Top}\text{-}K_i(p_t)$ with
$K_i=10$. Equal multiplicative weighting avoids a free mixing
hyperparameter and prevents either signal from dominating.

\paragraph{Token-to-token relation.}
For anchor rewriting we use the encoder's own symmetrized
self-attention,
\begin{equation}
s_{j,k} = \frac{1}{H}\sum_{h=1}^{H}
          \frac{A_t^{(h)}(j,k)+A_t^{(h)}(k,j)}{2},
\label{eq:relation}
\end{equation}
which captures \emph{functional} relatedness inside the encoder
rather than mere visual look-alikeness between patches.

\paragraph{Anchor rewriting.}
For each anchor $j\in I_t^{\text{idx}}$, we form a row-wise softmax
over $s_{j,\cdot}$ with the self-edge masked,
\begin{equation}
w_{j,k} = \begin{cases}
0 & k=j,\\[2pt]
\dfrac{\exp(s_{j,k})}{\sum_{m\neq j}\exp(s_{j,m})} & k\neq j,
\end{cases}
\label{eq:rewrite-weight}
\end{equation}
and produce the rewritten anchor by a self/neighbor blend,
\begin{equation}
\tilde v_{t,j}
= \rho\, v_{t,j} + (1-\rho)\!\sum_{k\neq j} w_{j,k}\, v_{t,k},
\qquad \rho = 0.6.
\label{eq:rewrite}
\end{equation}
The blend is taken over the whole image rather than a local
neighborhood, and $w_{j,k}$ is deliberately kept task-free: injecting
$\log p_t$ or $q_t$ into the softmax was found to collapse all $K_i$
anchors toward the same high-score tokens. The informative set is
\[
I_t = \{\tilde v_{t,j}\mid j\in I_t^{\text{idx}}\},\qquad |I_t| = K_i.
\]

\subsection{Teacher sequence}
\label{app:teacher-sequence}
The teacher hypothesis set for step $t$ is serialized as
\[
T_t = [G_t;\, I_t]\in\mathbb{R}^{N\times d_h},\qquad N = K_g + K_i = 20.
\]
The $K_g$ coverage tokens are placed before the $K_i$ informative
ones, a coarse-to-specific ordering that matches a causal decoder's
left-to-right view of the latent tokens.

\subsection{Latent-placeholder assembly}
\label{app:placeholder}
At every reasoning-step position, we insert $N=20$ special latent
placeholder tokens between the learned boundary markers,
\[
\langle\texttt{lat\_start}\rangle,\;\texttt{PAD}_1,\ldots,\texttt{PAD}_N,\;
\langle\texttt{lat\_end}\rangle.
\]
The number of placeholders is fixed at training time, so all $K_g$
coverage targets and all $K_i$ informative targets are realized
teacher targets; the latent objective reduces to a deterministic
cosine over a known target set without two-level masking. At
inference and at the RL stage we use $K_{\text{eval}}=8$ slots per
step as a common evaluation budget. An inference-time sweep from 4 to
20 shows a broad performance plateau (Appendix~\ref{app:slot-sweep}).

\subsection{Norm-aligned teacher injection}
\label{app:norm-align}
Teacher targets $\tau_{t,i}\in T_t$ live in the vision-encoder
post-merger space, which shares dimensionality with the LLM input
embedding space but not its scale. Let
$\mu_e = \mathbb{E}_w[\|e_w\|_2]$ denote the mean $L_2$ norm of the
LLM input embedding matrix, computed once at initialization. Each
teacher vector is rescaled in a direction-preserving way,
\begin{equation}
\tau_{t,i}^{\text{aligned}}
= \mu_e \cdot \frac{\tau_{t,i}}{\|\tau_{t,i}\|_2 + \varepsilon},
\label{eq:tau-aligned}
\end{equation}
and replaces the input embedding at the corresponding latent
placeholder. The boundary-marker embeddings are kept at their learned
values.

\subsection{EMA teacher}
\label{app:ema}
Teacher construction requires $h_t$, which itself co-evolves with the
student during training. To decouple the supervision signal from
high-frequency student updates, we maintain an exponential moving
average of the student parameters, following the mean-teacher framework~\citep{NIPS2017_68053af2}
\begin{equation}
\theta_{\text{tea}} \leftarrow \tau\,\theta_{\text{tea}}
+ (1-\tau)\,\theta_{\text{stu}},\qquad \tau = 0.999.
\label{eq:ema}
\end{equation}
At every step a single EMA forward (with the visual encoder frozen)
produces $h_t$, $V_t$, and $\{A_t^{(h)}\}$ in one pass; teacher
targets $T_t$ are then constructed from these.

\subsection{Stage-1 objective}
\label{app:stage1}
Let $\mathcal{S}_{\text{text}}$ and $\mathcal{S}_{\text{marker}}$
denote the sequence positions of ordinary text tokens and boundary
markers, respectively. The text objective is
\begin{equation}
\mathcal{L}_{\text{text}}
= -\!\!\!\!\sum_{\ell\in
        \mathcal{S}_{\text{text}}\cup\mathcal{S}_{\text{marker}}}\!\!\!\!
  \log p_\theta(y_\ell\mid y_{<\ell}, x, q),
\label{eq:lce}
\end{equation}
with labels at latent and image positions set to the ignore index;
the boundary markers remain supervised so that the model learns when
to open and close the latent tokens.

Let $\mathcal{S}_{\text{lat}}$ collect the $(t,i)$ pairs indexing
every latent slot in the sequence, and let $z_{t,i-1}$ denote the
student's last-layer hidden state at the token immediately preceding
the $i$-th slot of step $t$ (for $i=1$ this is the position of
$\langle\texttt{lat\_start}\rangle$). The shifted-cosine latent
objective is
\begin{equation}
\mathcal{L}_{\text{lat}}
= \frac{1}{|\mathcal{S}_{\text{lat}}|}\!\!
  \sum_{(t,i)\in\mathcal{S}_{\text{lat}}}\!\!
  \bigl(1 - \cos(z_{t,i-1},\, \tau_{t,i}^{\text{aligned}})\bigr).
\label{eq:lat-shifted}
\end{equation}
Direction alignment is the only degree of freedom because
\eqref{eq:tau-aligned} already matches norms.

The Stage-1 loss combines the two with a linear warm-up on the
latent weight,
\begin{equation}
\mathcal{L}_{\text{Stage1}}
= \mathcal{L}_{\text{text}} + \lambda_{\text{lat}}(s)\,
                              \mathcal{L}_{\text{lat}},
\label{eq:stage1}
\end{equation}
\begin{equation}
\lambda_{\text{lat}}(s)
= \min\!\bigl(s/0.1\,S_{\text{total}},\; 1\bigr)\,
  \lambda_{\text{lat}}^{\star},
\label{eq:warmup}
\end{equation}
where $s$ is the optimization step. The warm-up prevents
early-training cosine gradients from overwhelming
$\mathcal{L}_{\text{text}}$ while the student has not yet learned to
use the latent placeholders. In implementation, the alignment in
\eqref{eq:lat-shifted} corresponds to comparing
\texttt{outputs.hidden\_states[..., :-1, :]} with the assembled
\texttt{inputs\_embeds[..., 1:, :]} under the latent-placeholder
mask.

\subsection{Stage-2 objective}
\label{app:stage2}
\paragraph{Rollout.}
For each prompt $(x,q)$ we draw $G$ rollouts from the Stage-1 policy
$\pi_{\theta_{\text{old}}}$. Each rollout $g$ consists of a token
sequence $\mathbf{y}^{(g)}$ (interleaving text and
$\langle\texttt{lat\_pad}\rangle$ placeholders) and the latent hidden
states $\mathbf{z}^{(g)} = \{z^{(g)}_{t,i}\}_{(t,i)\in
\mathcal{S}_{\text{lat}}}$ emitted at every latent slot. Latent states
are captured directly from the decoder loop.

\paragraph{Reward and advantage.}
Each rollout receives a scalar reward
\begin{equation}
R_g
= \alpha_o\,\mathbb{1}[\text{answer}_g{=}\text{gold}]
+ \alpha_f\,\mathbb{1}[\text{format}_g\text{ valid}],
\label{eq:reward}
\end{equation}
combining outcome correctness with a light format term. The advantage
is the group-relative normalization
\begin{equation}
\hat A_g
= \frac{R_g - \operatorname{mean}(R_{1:G})}
        {\operatorname{std}(R_{1:G}) + \epsilon}.
\label{eq:advantage}
\end{equation}

\paragraph{Text policy term.}
On text positions we apply the standard clipped PPO objective to
token log-probabilities,
\begin{equation}
\begin{split}
\mathcal{L}_{\text{text}}^{\text{RL}}
&= -\,\mathbb{E}_{g,\ell}\bigl[
   \min\!\bigl(r_\ell^{(g)}\,\hat A_g,\\
&\qquad \operatorname{clip}(r_\ell^{(g)}, 1{-}\varepsilon, 1{+}\varepsilon)\,
              \hat A_g\bigr)\bigr],
\end{split}
\label{eq:text-rl}
\end{equation}
with
$r_\ell^{(g)} = \pi_\theta(y_\ell^{(g)}\mid y_{<\ell}^{(g)}) /
                \pi_{\theta_{\text{old}}}(y_\ell^{(g)}\mid y_{<\ell}^{(g)})$.
\mvh{} introduces no modification to text-token optimization.

\paragraph{Advantage-signed latent alignment.}
Latent slots have no discrete log-probability. We re-inject each
rollout's captured latent states $z^{(g)}_{t,i}$ (detached, no
gradient) at the slot positions of $\mathbf{y}^{(g)}$ and run a
training forward pass under $\theta$ to obtain on-policy slot states
$z^{\theta}_{t,i}$. Advantage is then propagated through cosine
alignment,
\begin{equation}
\mathcal{L}_{\text{lat}}^{\text{RL}}
= -\,\mathbb{E}_{g,(t,i)}\!\bigl[
   \hat A_g \cdot \cos\!\bigl(z^{\theta}_{t,i},\, z^{(g)}_{t,i}\bigr)
\bigr],
\label{eq:lat-rl}
\end{equation}
so that positive-advantage rollouts pull the policy toward the
rollout trajectory and negative-advantage rollouts push it away,
mirroring the reinforce/suppress semantics of the text-token PPO
term.

\paragraph{Latent trust region.}
Equation~\eqref{eq:lat-rl} alone is direction-unconstrained for
negative advantages. We anchor the policy to the frozen Stage-1
checkpoint $\theta_{\text{SFT}}$: a reference forward pass under
$\theta_{\text{SFT}}$ on the same rollout sequence yields anchor
states $z^{\text{SFT}}_{t,i}$, and we add
\begin{equation}
\mathcal{L}_{\text{anchor}}^{\text{RL}}
= \mathbb{E}_{g,(t,i)}\!\bigl[
   1 - \cos\!\bigl(z^{\theta}_{t,i},\, z^{\text{SFT}}_{t,i}\bigr)\bigr].
\label{eq:anchor-rl}
\end{equation}
This plays the continuous-space role that PPO's reference-KL plays
on the text distribution: latent updates are bounded to a
direction-only neighborhood of the MVH-aligned Stage-1 manifold.

\paragraph{Full Stage-2 loss.}
\begin{equation}
\mathcal{L}_{\text{Stage2}}
= \mathcal{L}_{\text{text}}^{\text{RL}}
+ \lambda_{\text{lat}}^{\text{RL}}\,\mathcal{L}_{\text{lat}}^{\text{RL}}
+ \beta_{\text{anchor}}\,\mathcal{L}_{\text{anchor}}^{\text{RL}}.
\label{eq:stage2}
\end{equation}
The three terms act on disjoint aspects of the policy:
$\mathcal{L}_{\text{text}}^{\text{RL}}$ keeps text-token selection
inside the standard PPO framework;
$\mathcal{L}_{\text{lat}}^{\text{RL}}$ steers the continuous latent
representation under the same GRPO advantage; and
$\mathcal{L}_{\text{anchor}}^{\text{RL}}$ bounds the latent update
within a trust region around the Stage-1 checkpoint. All weights and
clip parameters are listed in Table~\ref{tab:hyperparams}.


\begin{table}[h]
\centering
\footnotesize
\setlength{\tabcolsep}{3pt}
\input{tables/tabA1_hyperparams}
\caption{Core \mvh{} hyperparameters used in the experiments. The
query temperature $\tau_q=d_h^{-1/2}$ and the rescaling target
$\mu_e$ are design constants rather than tunable hyperparameters.}
\label{tab:hyperparams}
\end{table}

\begin{table*}[t]
\centering
\small
\setlength{\tabcolsep}{3pt}
\resizebox{\textwidth}{!}{\input{tables/tabA2_full_accuracy}}
\caption{\textbf{Full per-benchmark accuracy including sub-category scores and additional comparisons.} Settings analyzed in this paper are in the lower block. Gray rows mark reasoning-with-tool methods. The \textbf{bold numbers} indicate the best performance achieved by each setting, and the \uline{underline numbers} are the second best. PixelReasoner and DeepEyes are excluded from this ranking. Stars ($^{*}$) indicate values reproduced from prior reports; dashes indicate scores not reported on the corresponding benchmark.}
\label{tab:full-accuracy}
\end{table*}

\section{Reproduction Details and Full Accuracy}
\label{app:training-data}

\paragraph{Training data.}
The SFT corpus used by \mvh-SFT, ILVR-Stage1, and ILVR-Stage2 is taken from Monet's released SFT data, which keeps the SFT comparison fully matched. Each sample interleaves text reasoning steps with helper images such as cropped regions or annotated views; during training we replace each helper image with $N=20$ latent placeholder tokens and build the teacher targets from the original image, as described in Appendix~\ref{sec:method-appendix}. Monet's RL corpus is not public, so we follow its construction pipeline on DeepEyes-v2 to assemble an RL set for \mvh-RL. The original RL-stage comparison between \mvh-RL and Monet-RL is therefore matched in pipeline rather than in exact data. As an additional control, we retrain Monet with its original VLPO objective on the exact \mvh-RL corpus (Appendix~\ref{app:monet-matched}). Monet-SFT and Monet-RL checkpoints otherwise come from the original authors and are used without modification. Detailed hyperparameter settings are summarized in Table~\ref{tab:hyperparams}.

\paragraph{Evaluation protocol.}
Evaluation runs through VLMEvalKit~\citep{duan2025vlmevalkitopensourcetoolkitevaluating} on V$^*$, HRBench-4K, HRBench-8K, and MME-RealWorld-Lite. All six method-stage settings share an inference latent budget of $K_{\text{eval}}=8$ slots per step. An inference-time sweep from 4 to 20 reveals a broad performance plateau (Appendix~\ref{app:slot-sweep}). Scoring follows each benchmark's official multiple-choice protocol, and per-sample correctness is the unit used by the marker-only coverage (Appendix~\ref{app:moc}) and the cross-example swap (Appendix~\ref{app:slot-swap}) analyses below. The benchmark-averaged numbers cited in the main text are unweighted means over the four benchmarks. The original six-setting accuracy results are averaged over multiple independent runs; confidence intervals for the sample-level extension analyses use 1,000 bootstrap resamples.

\paragraph{Full per-benchmark accuracy.}
Table~\ref{tab:full-accuracy} gives the full version of Table~\ref{tab:accuracy-setup} from the main text, adding sub-category scores (Attribute/Spatial on V$^*$, FSP/FCP on HRBench, Reasoning/Perception on MME-RealWorld-Lite) and comparisons against GPT-4o, DeepEyes~\citep{zheng2026deepeyes}, Pixel Reasoner~\citep{NEURIPS2025_0c38f547}, and LVR. We follow Monet's training setup for this table: \mvh-SFT runs for 3 epochs and \mvh-RL for 2 epochs, and ILVR-Stage1 and ILVR-Stage2 each run for 3 epochs.

\paragraph{Fine-grained perception (CoMT).}
Table~\ref{tab:comt} reports results on CoMT~\citep{cheng2025comt}, a fine-grained perception benchmark with four sub-tasks: creation, deletion, selection, and update. We train \mvh{} on the CoMT training data released by ILVR, and the \mvh-SFT stage follows the same training setup as ILVR-Stage1 to keep the two SFT-stage methods directly comparable. Following ILVR's training setup, \mvh-SFT and ILVR-Stage1 each run for 15 epochs, and \mvh-RL and ILVR-Stage2 each run for 2 epochs. Results are reported on both Qwen2.5-VL-7B and Qwen3-VL-8B backbones.

\begin{table*}[t]
\centering
\footnotesize
\setlength{\tabcolsep}{3pt}
\input{tables/tabA3_comt}
\caption{\textbf{CoMT benchmark accuracy across Qwen2.5-VL-7B and Qwen3-VL-8B backbones.} Four sub-tasks (creation, deletion, selection, update) are evaluated separately and as an unweighted average. The \textbf{bold numbers} indicate the best performance achieved by each setting, and the \uline{underline numbers} are the second best.}
\label{tab:comt}
\end{table*}

\section{Additional Controls and Evaluation Scope}
\label{app:additional-evaluation}

The main diagnostic suite covers the six settings in
Section~\ref{sec:settings-diagnostics}. The additional evaluations
broaden selected tests rather than repeating every diagnostic for every
checkpoint. Table~\ref{tab:diagnostic-coverage} records the scope of
each extension.

\begin{table*}[t]
\centering
\small
\setlength{\tabcolsep}{5pt}
\input{tables/tabA4_diagnostic_coverage}
\caption{\textbf{Coverage of the diagnostic suite and additional
evaluations.} The original six settings receive the full diagnostic
suite. Mirage and LVR extend the core slot/marker tests; TextVQA and the
causal attention intervention have the narrower scopes shown here.}
\label{tab:diagnostic-coverage}
\end{table*}

\subsection{Additional Checkpoints}
\label{app:generality}

We extend the core slot/marker analysis to Mirage-Stage1,
Mirage-Stage2, and LVR-SFT. Table~\ref{tab:generality-accuracy} places
the three checkpoints on the same four benchmarks, and
Table~\ref{tab:generality-diagnostics} reports slot-content
perturbation, marker sensitivity, and marker-only coverage.

\begin{table}[t]
\centering
\footnotesize
\setlength{\tabcolsep}{2pt}
\input{tables/tabA5_generality_accuracy}
\caption{\textbf{Accuracy of the additional checkpoints (\%).}
Avg. is the unweighted mean over the four multiple-choice benchmarks.}
\label{tab:generality-accuracy}
\end{table}

\begin{table*}[t]
\centering
\small
\setlength{\tabcolsep}{3pt}
\resizebox{\textwidth}{!}{\input{tables/tabA6_generality_diagnostics}}
\caption{\textbf{Core diagnostics for the additional checkpoints
(percentage points and \%).} For these checkpoints, Slot-sens. is
$\text{Normal}-\operatorname{avg}(\text{ZeroSlot},\text{RandomSlot})$;
\textsc{FixedSlot} was not run. Marker-sens. is
$\text{Normal}-\text{ZeroMarker}$. MOC brackets are 95\% sample-level
bootstrap confidence intervals ($B=1000$).}
\label{tab:generality-diagnostics}
\end{table*}

Slot-content sensitivity is at most $0.3$ points for the three added
checkpoints, while marker sensitivity ranges from $0.2$ to $62.6$
points. Mirage's Stage1-to-Stage2 change also reproduces the shift from
strong to weak marker dependence. Thus the core pattern of weak
slot-value sensitivity and heterogeneous dependence on markers and format holds
across five methods and nine checkpoints; we do not assume that the
narrower diagnostics in Table~\ref{tab:diagnostic-coverage} extend to
all nine.

\subsection{Open-Ended TextVQA Extension}
\label{app:textvqa}

We additionally evaluate \mvh-SFT and \mvh-RL on the TextVQA~\citep{singh2019towards}
validation split. Normal and intervention columns in
Table~\ref{tab:textvqa} use the official continuous
\texttt{vqa\_score}; a threshold of $0.5$ is used only to construct the
binary solved sets required by MOC. Slot-sens. again averages the two
available perturbations because \textsc{FixedSlot} was not run.

\begin{table*}[t]
\centering
\small
\setlength{\tabcolsep}{2pt}
\resizebox{\textwidth}{!}{\input{tables/tabA7_textvqa}}
\caption{\textbf{Core diagnostics on open-ended TextVQA (\%).}
$K=0$ retains the two boundary markers while removing the slots;
\textsc{NoLatent} is used only to define the MOC denominator.
$n_{\mathrm{gain}}$ is the number of examples in the latent-gained
set.}
\label{tab:textvqa}
\end{table*}

Perturbing slot contents leaves the continuous score essentially flat,
while marker-only decoding recovers $99.6\%$ and $92.7\%$ of the
latent-gained sets for \mvh-SFT and \mvh-RL, respectively. This is an
open-ended extension of the core pattern, not a fifth benchmark for the
full diagnostic suite.

\subsection{Matched-Data Monet-RL Control}
\label{app:monet-matched}

To separate RL-data composition from the comparison in
Section~\ref{sec:settings-diagnostics}, we start from Monet's released
SFT checkpoint and run its original VLPO objective on the exact RL
corpus used by \mvh-RL. Table~\ref{tab:monet-matched} shows that exact
data matching leaves Monet's average accuracy essentially unchanged
($67.5$ versus $67.6$). RL-data composition is therefore unlikely to
be the main explanation for the observed gap. The methods still differ
in SFT supervision, latent-target construction, and RL objective, so
this control does not isolate architecture.

\begin{table}[t]
\centering
\footnotesize
\setlength{\tabcolsep}{2pt}
\input{tables/tabA8_monet_matched}
\caption{\textbf{Exact RL-data control for Monet (accuracy, \%).}
Monet-RL-matched uses Monet's SFT checkpoint and VLPO objective with the
exact RL corpus used by \mvh-RL.}
\label{tab:monet-matched}
\end{table}

\subsection{Inference-Time Slot-Count Sweep}
\label{app:slot-sweep}

Stage-1 trains with $N=20$ targets, whereas the RL stage and the main
evaluation use $K_{\mathrm{eval}}=8$. We vary the active inference
budget from 4 to 20 on the fixed \mvh-SFT and \mvh-RL checkpoints.
Table~\ref{tab:slot-sweep} shows a broad plateau: benchmark-averaged
accuracy varies by at most $0.7$ points for either checkpoint. The
sweep does not retrain a separate model at each slot count and therefore
does not remove the Stage-1 train/inference mismatch.

\begin{table}[t]
\centering
\footnotesize
\setlength{\tabcolsep}{2pt}
\resizebox{\columnwidth}{!}{\input{tables/tabA9_slot_sweep}}
\caption{\textbf{Inference-time slot-count sweep (benchmark-averaged
accuracy, \%).} Each value is evaluated from a fixed checkpoint without
retraining.}
\label{tab:slot-sweep}
\end{table}

\section{Ablation Studies}
\label{app:ablations}

We perform two ablations on \mvh{}'s design choices. The first targets the rewriting blend weight $\rho$ used in anchor rewriting, with all variants trained from a Qwen2.5-VL-7B initialization. $\rho=0.6$ achieves the best benchmark-averaged accuracy (Table~\ref{tab:ablation-rho}). The second targets the latent RL design, with all variants initialized from the \mvh-SFT checkpoint. Our advantage-signed latent alignment outperforms applying vanilla GRPO directly to the latent positions (Table~\ref{tab:ablation-rl}).

\begin{table}[t]
\centering
\footnotesize
\setlength{\tabcolsep}{2.8pt}
\input{tables/tabA10_ablation_rho}
\caption{Sensitivity to the rewriting blend weight $\rho$. Benchmark-averaged accuracy on V$^*$, HRBench-4K, HRBench-8K, and MME-RealWorld-Lite. The highlighted row is the \mvh-SFT setting. The \textbf{bold numbers} indicate the best performance achieved by each MLLM, and the \uline{underline numbers} are the second best.}
\label{tab:ablation-rho}
\end{table}

\begin{table}[t]
\centering
\footnotesize
\setlength{\tabcolsep}{3pt}
\resizebox{\columnwidth}{!}{\input{tables/tabA11_ablation_rl}}
\caption{Comparison between vanilla GRPO and our advantage-signed latent alignment on the same SFT checkpoint. Benchmark-averaged accuracy on V$^*$, HRBench-4K, HRBench-8K, and MME-RealWorld-Lite. The \textbf{bold numbers} indicate the best performance achieved by each setting.}
\label{tab:ablation-rl}
\end{table}

\section{Structural Sanity Diagnostics}
\label{app:structural}

These diagnostics support the probe-legitimacy claim in \S\ref{sec:mvh-probe}: before running causal interventions, we verify that \mvh{} latents are nontrivial enough for the slot-content memory hypothesis to be worth testing. They are not intended to prove content causality.

\begin{figure*}[t]
\centering
\includegraphics[width=\textwidth]{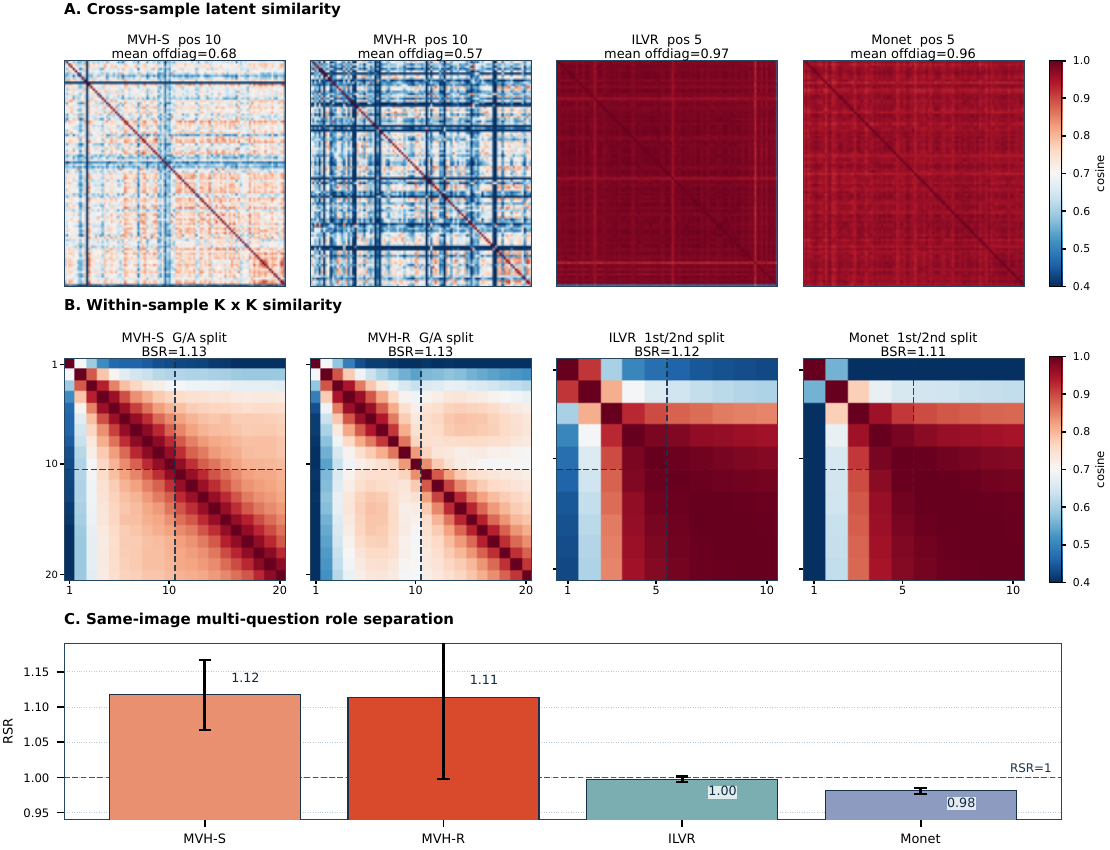}
\caption{\textbf{Structural diagnostics for latent organization.} (A) Cross-sample latent similarity shows that \mvh{} latents vary more across inputs than ILVR and Monet, which are closer to cross-sample collapse. (B) Within-sample $K\times K$ similarity recovers the intended coverage/anchor split in \mvh{}; for ILVR and Monet, the first/second-half split is only a positional reference, not a designed role split. (C) Same-image multi-question analysis quantifies role separation with RSR, the ratio between coverage-state and anchor-state similarity across questions. \mvh{} has RSR above 1.1, while ILVR and Monet remain near 1.0.}
\label{fig:app_structural_diagnostics}
\end{figure*}

\paragraph{Cross-sample latent diversity.}
For each setting we dump the latent hidden states on 100 V$^*$ samples. At every latent slot index $k$, we then compute the $100\times 100$ cosine-similarity matrix of the slot-$k$ states across samples. A latent that varies with the input produces visible off-diagonal structure and contrast between diagonal and off-diagonal entries; a latent that collapses to a near-constant representation produces a uniformly high matrix. \mvh-SFT, \mvh-RL and ILVR-Stage1 show strong cross-sample diversity at every slot index, while ILVR-Stage2, Monet-SFT, and Monet-RL stay close to a uniform matrix, consistent with cross-sample collapse.

\paragraph{Within-sample block structure.}
For each sample we compute the $K\times K$ cosine matrix between its $K$ latent slots and average this matrix across 100 samples. Under the \mvh{} design $K{=}20$ slots are split into a coverage half and an informative half, and a $2\times 2$ block pattern is expected in this matrix (within-half similarity higher than cross-half similarity). The Block Sensitivity Ratio (BSR) is the ratio of mean within-block similarity to mean cross-block similarity; values above one indicate that the two halves have been learned as functionally distinct subsets. \mvh{} obtains $\text{BSR}\approx 1.13$, while ILVR and Monet do not have a designed block structure and their first/second-half split is reported only as a positional reference.

\paragraph{Role separation across questions.}
The within-sample analysis cannot distinguish coverage from informative slots by function; it only checks that the two halves are not identical. To target the designed role separation, we take 20 images with three to five different questions each, dump latents for every (image, question) pair, and compute the cross-question similarity inside the coverage half and inside the informative half separately. The coverage half is expected to be query-independent (high cross-question similarity) and the informative half query-dependent (lower cross-question similarity); the Role Separation Ratio (RSR) is their ratio. \mvh{} reaches $\text{RSR}\approx 1.12$ on this set, while ILVR and Monet remain near $1.00$ since they have no designed role split.

\section{Intervention Suite}
\label{app:interventions}

\paragraph{Intervention conditions.}
Table~\ref{tab:app_intervention_conditions} summarizes the latent-interface interventions used throughout the paper. The conditions are designed to separate four factors: the presence of the latent format, the precise contents of latent slots, the identity of learned boundary markers, and malformed execution when a corrupted span remains in the sequence.

\begin{table*}[t]
\centering
\small
\setlength{\tabcolsep}{3pt}
\input{tables/tabA12_intervention_conditions}
\caption{\textbf{Latent-interface intervention conditions.} The intervention suite separates clean latent removal, slot-content perturbation, boundary-marker corruption, and bare-slot stress tests. $K=0$ keeps learned boundary markers while removing latent slots; \textsc{NoLatent} removes the span cleanly; \textsc{ZeroMarker} zeroes marker input embeddings while keeping the corrupted marker positions in the sequence. Forced pads without markers are treated as an out-of-distribution stress test.}
\label{tab:app_intervention_conditions}
\end{table*}

\paragraph{Implementation of interventions.}
All interventions are applied at the input-embedding level, after token assembly but before the first decoder block. $K{=}0$ keeps the boundary-marker positions and drops the latent placeholders entirely, so the model decodes from $\langle\texttt{lat\_start}\rangle$ immediately followed by $\langle\texttt{lat\_end}\rangle$. \textsc{NoLatent} additionally bans the boundary markers through a logits processor, producing a clean removal of the latent format. \textsc{ZeroSlot}, \textsc{RandomSlot}, and \textsc{FixedSlot} replace the slot input embeddings with zeros, Gaussian noise at the matching scale, or a position-conditioned fixed vector while leaving the marker positions and sequence layout untouched. \textsc{ZeroMarker} zeros the marker input embeddings while keeping the marker positions in place, which lets the corrupted span propagate through generation. \textsc{ZeroAll} combines slot and marker zeroing. The forced-pad-without-markers condition inserts a bare slot span with no enclosing markers and is treated as an out-of-distribution stress test rather than a clean comparison.

\paragraph{Sensitivity metrics.}
We derive four scalar sensitivities from the seven principal conditions, each of which condenses a row of the intervention matrix into a one-row-per-setting summary. Slot-content sensitivity is $\text{Normal} -
\operatorname{avg}(\text{ZeroSlot}, \text{RandomSlot}, \text{FixedSlot})$, which is close to zero in every setting. Marker sensitivity is $\text{Normal}-\text{ZeroMarker}$, and it separates the three marker-dependence ranges discussed in \S\ref{sec:regimes}. Latent-format dependence is $\text{Normal}-\text{NoLatent}$, which captures how much accuracy depends on the latent tokens as a whole. Malformed interference is $\text{NoLatent}-\text{ZeroMarker}$, which compares the cost of keeping a corrupted span in the sequence against the cost of removing the span cleanly; a positive value indicates that the model has a learned protocol for the latent tokens and is hurt more by a corrupted version than by a clean removal.

\begin{figure*}[t]
\centering
\includegraphics[width=\textwidth]{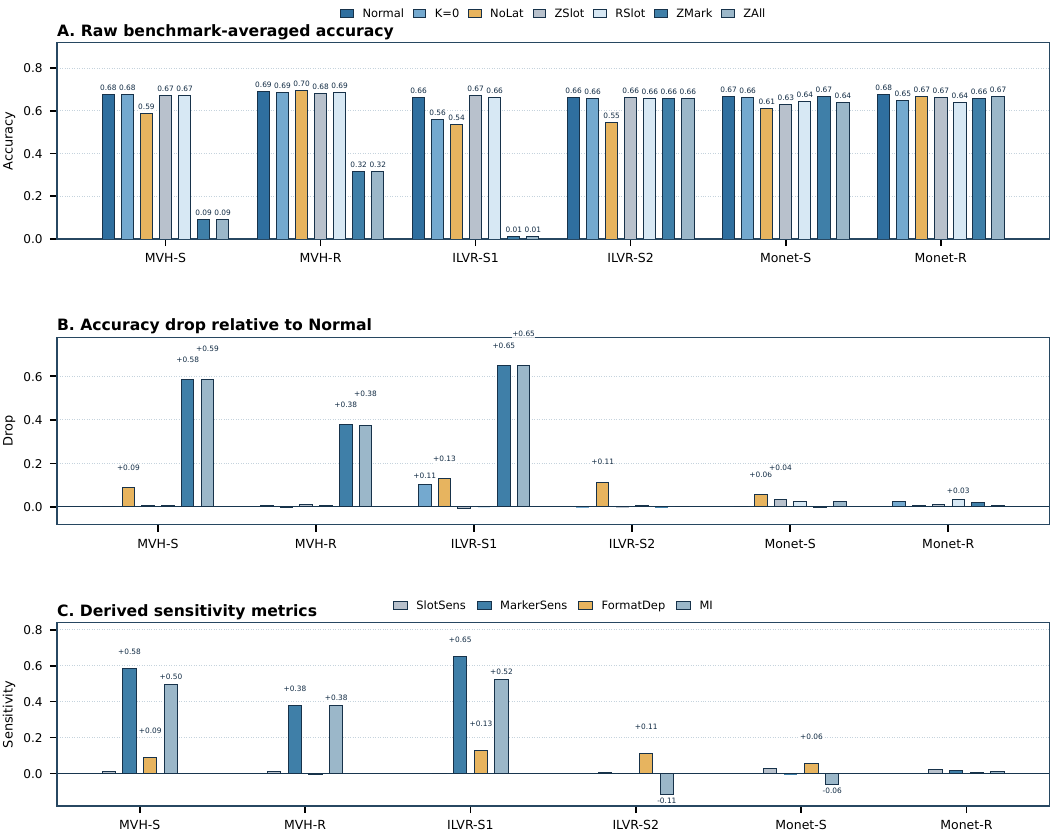}
\caption{\textbf{Full latent-interface intervention matrices.} (A) Benchmark-averaged raw accuracy under each intervention condition. (B) Accuracy drop relative to the Normal condition. (C) Derived sensitivity metrics used to classify marker-dependence ranges. Slot-content perturbations remain close to Normal, whereas \textsc{ZeroMarker} and \textsc{ZeroAll} selectively collapse MVH-SFT, MVH-RL, and ILVR-Stage1.}
\label{fig:app_full_intervention_matrix}
\end{figure*}

\paragraph{Full intervention matrix.}
Figure~\ref{fig:app_full_intervention_matrix} expands Figure~\ref{fig:marker-protocol} with raw benchmark-averaged accuracy, accuracy drops relative to Normal, and the derived sensitivity metrics used to define the marker-dependence ranges.

\section{Slot-Content Memory Tests: Detailed Results}
\label{app:content-tests}

This appendix expands the four tests of slot-content memory reported in \S\ref{sec:pad-tests}, following the order of tests in \S\ref{sec:four-tests}. Each test uses a different probing assumption: a static perturbation test on slot contents, an image-free injection test, a cross-example swap test, and a sample-level marker-only coverage test. The subsections below give the protocol and the data behind the corresponding row of the consolidated content-tests table.

\subsection{Overview}
Table~\ref{tab:content-tests} consolidates the four tests into a single predicted-versus-observed view.

\begin{table*}[t]
\centering
\small
\setlength{\tabcolsep}{3pt}
\input{tables/tabA13_content_tests}
\caption{Tests of the visual-memory account. Even under content-favorable \mvh{} supervision, slot contents do not behave like recoverable visual memories.}
\label{tab:content-tests}
\end{table*}

\subsection{Slot-Value Perturbation}
This test asks whether the contents of the slots are causally responsible for the latent tokens' gain by replacing them with substitutes that retain the sequence layout. \textsc{ZeroSlot} sets the slot input embeddings to zero, \textsc{RandomSlot} samples Gaussian noise at the matching scale, and \textsc{FixedSlot} reuses a position-conditioned vector that is independent of the input. The benchmark-averaged accuracy under these three conditions is reported in Figure~\ref{fig:app_full_intervention_matrix}(A) as part of the full intervention matrix. Across the six settings, perturbing slot contents moves accuracy by at most $0.027$, while the same models respond strongly to interventions on the boundary markers (\S\ref{app:marker-controls}). The slot-content sensitivity metric defined in \S\ref{app:interventions} collapses this into a single number per setting, which stays near zero in every case.

\begin{table}[t]
\centering
\footnotesize
\setlength{\tabcolsep}{3pt}
\input{tables/tabA14_perturbation_robustness}
\caption{\textbf{Perturbation robustness across the original six
settings (percentage points).} Slot-sens. is the four-benchmark mean of
$\text{Normal}-\operatorname{avg}(\text{ZeroSlot},\text{RandomSlot},
\text{FixedSlot})$; brackets give the range across benchmarks.
Marker-sens. is $\text{Normal}-\text{ZeroMarker}$.}
\label{tab:perturbation-robustness}
\end{table}

\subsection{Image-Free Injection}
\label{app:blind-injection}

As a stronger portability test, we remove the image and ask whether latent states extracted from an image-conditioned run can restore the missing visual information. Figure~\ref{fig:causal-tests}B reports the aggregate comparison across the six settings; the paired uncertainty analysis below targets \mvh-SFT, where the Stage-1 teacher provides direct visual-space supervision for the slot states.

The probe runs as a paired three-condition test on the V$^*$ subset that emits valid latent tokens. We first run \mvh-SFT with the full image and capture the final-layer hidden state at every emitted latent slot ($n=58$ samples). We then re-run the same prompt with the image replaced by zeros, both with the captured slot states injected at the corresponding positions ($n=58$) and without any injection ($n=57$ after a prompt-formatting filter). If the captured states encode the visual evidence that the answer needs, blind accuracy with injection should rise toward the full-image level.

\begin{table}[t]
\centering
\small
\setlength{\tabcolsep}{3pt}
\input{tables/tabA15_blind_injection}
\caption{\textbf{Blind-run latent injection test for \mvh-SFT on V$^*$.} Image-conditioned latent states are extracted from a full-image run and injected into a blind run. Injected latents improve accuracy from 0.228 to 0.276, yielding a recovery rate of $(A_{\mathrm{inj}}-A_{\mathrm{blind}})/(A_{\mathrm{full}}-A_{\mathrm{blind}})=0.078$ on the valid subset ($n=58$; 95\% bootstrap CI $[-15.2, 28.6]$).}
\label{tab:app_blind_injection}
\end{table}

Table~\ref{tab:app_blind_injection} shows that injection recovers only a small fraction of the full-image gap. Its wide interval includes zero, so this result is directionally consistent with the other tests but is not decisive on its own. It does not imply that the latent tokens are useless; rather, it provides limited evidence that the image-conditioned slot states are not portable enough to restore visual performance when the image signal is removed.

\begin{figure*}[t]
\centering
\includegraphics[width=\textwidth]{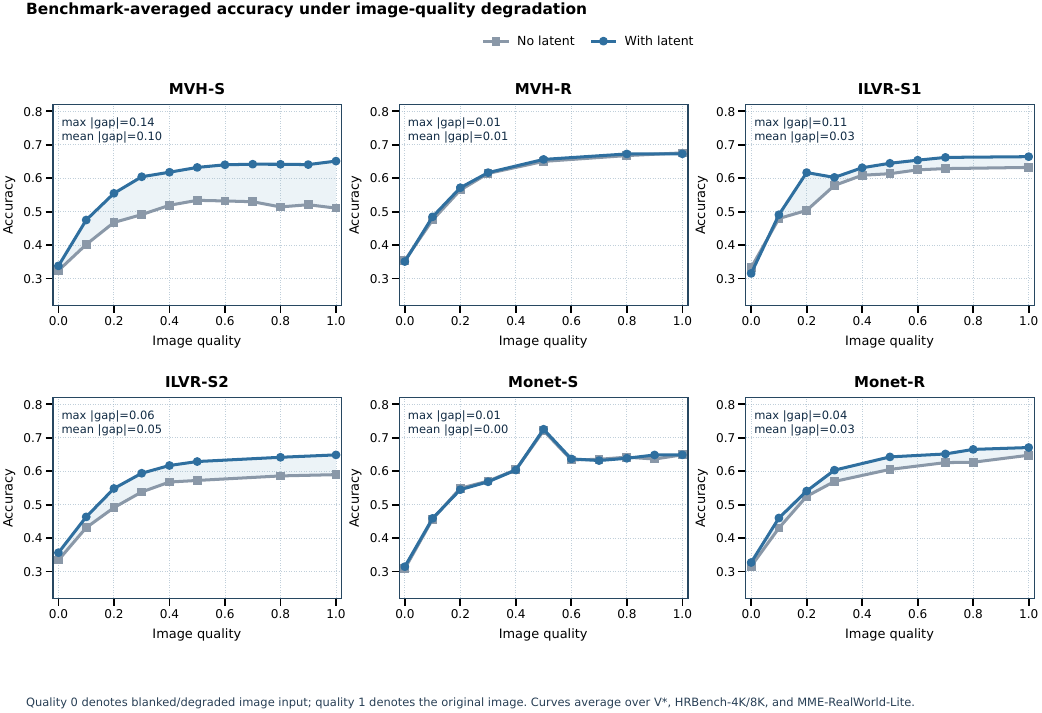}
\caption{\textbf{Image-quality bottleneck curves.} Accuracy is measured as image quality is progressively degraded from blanked input ($0.0$) to the original image ($1.0$). Curves report paired benchmark averages over V$^*$, HRBench-4K/8K, and MME-RealWorld-Lite for with-latent and no-latent decoding. The curves show no consistent mid-quality kick-in pattern: some settings have a persistent offset, while others nearly overlap, but the latent tokens do not behave like an independent visual buffer that selectively compensates for missing image information.}
\label{fig:app_pixel_bottleneck}
\end{figure*}

\paragraph{Image-quality bottleneck.}
As a continuous-degradation version of the same test, we degrade the image gradually and ask whether the latent tokens compensate for missing visual evidence. A standalone latent visual buffer would predict a selective separation between with-latent and no-latent runs at intermediate image quality. We sweep image quality across six levels (blanked at $0.0$, then $0.2$, $0.3$, $0.5$, $0.8$, and the original image at $1.0$), apply matched downsampling and blur at each level, and run paired with-latent and no-latent decodes on V$^*$, HRBench-4K, HRBench-8K, and MME-RealWorld-Lite for every setting (Figure~\ref{fig:app_pixel_bottleneck}).

\subsection{Cross-Example Slot Swap}
\label{app:slot-swap}

Zero, random, and fixed slot-content perturbations may be criticized as out-of-distribution corruptions. We therefore run an in-distribution swap test that uses real model executions as donors. We force a fixed latent prefix at the end of each prompt ($\langle\texttt{lat\_start}\rangle$, $K$ slot placeholders, $\langle\texttt{lat\_end}\rangle$), capture the slot hidden states at a chosen decoder layer during a clean forward pass, and re-inject the captured states at the same positions of a target sample. The protocol keeps marker positions, slot positions, and decoder layer identical across donor and target, so any change observed after the swap is attributable to the slot contents rather than to the layout. We test three decoder depths: an early layer (L5), a mid layer (L14), and a late layer (L27). Each (setting, layer) combination is evaluated under four donor conditions: a self-swap sanity check (target equals donor), a label-matched cross-sample swap (different image but the same multiple-choice answer letter), a label-mismatched cross-sample swap (different image and different answer), and a random donor baseline. The resulting grid is $6\text{ settings}\times 3\text{ layers}\times 4\text{ conditions}=72$ sub-jobs on the same $57$-sample V$^*$ subset used by the marker controls.

\begin{table}[t]
\centering
\small
\setlength{\tabcolsep}{4pt}
\input{tables/tabA16_latent_swap_sanity}
\caption{\textbf{Self-swap sanity check.} TrueMk+Slot denotes forced-prefix decoding with the trained boundary markers and the original slot states (no swap). $\Delta$ reports Self-swap@L14 minus the TrueMk+Slot accuracy. Replacing a target's slot states with its own captured states leaves forced-prefix accuracy unchanged up to small evaluation noise, validating the hook protocol before cross-sample donor swaps are interpreted.}
\label{tab:app_swap_sanity}
\end{table}

We additionally pool donor-answer adoption across V$^*$, HRBench-4K,
HRBench-8K, and MME-RealWorld-Lite. Table~\ref{tab:pooled-swap}
reports sample-level bootstrap intervals for this extension.

\begin{table}[t]
\centering
\small
\setlength{\tabcolsep}{5pt}
\input{tables/tabA17_pooled_swap}
\caption{\textbf{Donor-answer adoption pooled over all four
multiple-choice benchmarks.} Chance is 25\%; brackets are 95\%
sample-level bootstrap confidence intervals ($B=1000$). Every interval
lies below chance.}
\label{tab:pooled-swap}
\end{table}

\begin{figure*}[t]
\centering
\includegraphics[width=\textwidth]{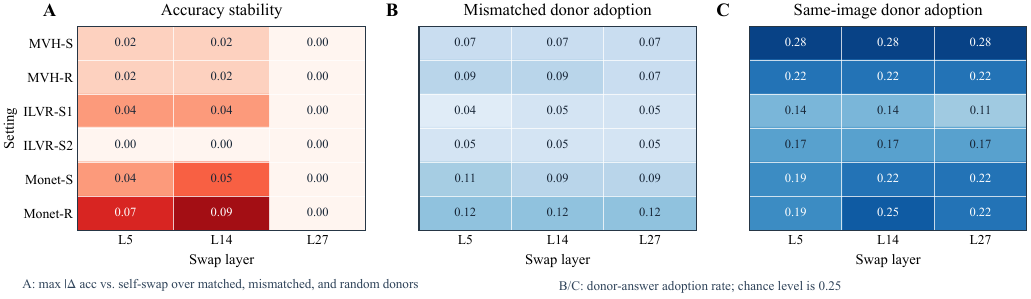}
\caption{\textbf{Full in-distribution latent-swap results.} We capture slot states from forced-prefix runs and replace the target slot states at the same decoder layer, using layers L5, L14, and L27. (A) Accuracy deviations from self-swap remain small across the main V$^*$ swap grid, indicating that real donor slot states do not substantially rewrite target predictions. (B) For label-mismatched donors, donor-answer adoption stays well below the 25\% chance level across settings and layers. (C) The same-image cross-question control asks whether slot states carry question-specific anchor information when the visual context is identical; donor-answer adoption remains near chance on the multi-question V$^*$ subset. Together, these results address the concern that slot-content perturbations are merely out-of-distribution corruptions: even in-distribution donor slot states do not behave as portable visual evidence.}
\label{fig:app_latent_swap}
\end{figure*}

Table~\ref{tab:app_swap_sanity} verifies that self-swap leaves forced-prefix accuracy unchanged, validating the hook protocol. Figure~\ref{fig:app_latent_swap} then shows that cross-sample donor states neither substantially move target accuracy nor transfer donor answers on the original V$^*$ analysis. In the four-benchmark extension, donor adoption remains $5.3\text{--}12.7\%$, with every interval below the 25\% chance rate (Table~\ref{tab:pooled-swap}). The same-image cross-question control is particularly diagnostic: even when donor and target share the same visual context, replacing slot states does not cause the decoder to adopt the donor question's answer.

\subsection{Marker-Only Coverage by Benchmark}
\label{app:moc}

Aggregate accuracy does not distinguish whether the same examples are solved across intervention conditions. We therefore report marker-only coverage at the sample level. Let $S_{\mathrm{Normal}}$ be the set of examples solved by normal latent decoding, $S_{K=0}$ the set solved when only the boundary markers are kept and the slots are dropped, and $S_{\mathrm{NoLatent}}$ the set solved when the entire latent tokens are removed. The latent tokens' unique solved set is $S_{\mathrm{Normal}} \setminus S_{\mathrm{NoLatent}}$, and marker-only coverage is the fraction of that set that is also solved by marker-only decoding:
\begin{equation}
\mathrm{MOC}
=
\frac{
|S_{K=0} \cap (S_{\mathrm{Normal}} \setminus S_{\mathrm{NoLatent}})|
}{
|S_{\mathrm{Normal}} \setminus S_{\mathrm{NoLatent}}|
}.
\label{eq:moc}
\end{equation}
$\mathrm{MOC}$ counts examples rather than averaging accuracy, so a high value means that the same examples uniquely solved by the latent tokens are also solved by a marker-only decode. A value close to one identifies settings in which the latent tokens' unique contribution is largely covered by the boundary markers; a value near zero would identify settings in which slot contents matter for the uniquely solved subset.

\begin{table*}[t]
\centering
\small
\setlength{\tabcolsep}{5pt}
\input{tables/tabA18_marker_covered_gain}
\caption{\textbf{Per-benchmark marker-only coverage.} Each cell reports $\mathrm{MOC}$ with $n_{\mathrm{inc}}=|S_{\mathrm{Normal}}\setminus S_{\mathrm{NoLatent}}|$ in parentheses. $\mathrm{MOC}$ is the fraction of the latent tokens' unique solved set that is also solved by marker-only decoding. The mean column is the unweighted average over benchmarks. \textsuperscript{$\dagger$} marks cells with $n_{\mathrm{inc}}<10$.}
\label{tab:app_mcg}
\end{table*}

Table~\ref{tab:app_mcg} expands the aggregate $\mathrm{MOC}$ values reported in Figure~\ref{fig:causal-tests}. ILVR-Stage1 reaches complete marker-only coverage on all four benchmarks, and \mvh{} remains high except for a lower but still substantial MME-RealWorld-Lite value for \mvh-SFT. Monet-SFT is weaker, especially on MME-RealWorld-Lite, while Monet-RL has small incremental sets in V$^*$ and MME-RealWorld-Lite. High $\mathrm{MOC}$ does not imply that latent slots are functionally irrelevant; it shows that the examples uniquely gained by the latent tokens are usually not dependent on precise slot contents.

\section{Boundary-Marker Controls}
\label{app:marker-controls}

The \kzero{} and marker-only coverage results show that learned boundary markers cover much of the latent-token gain. We next ask whether this is a generic effect of adding any special or pause-like tokens. To avoid ambiguity from free latent generation, we use forced-prefix decoding: a fixed prefix is appended to the prompt, and the model decodes only after that prefix.

Nine prefix variants are compared on the same 57-sample V$^*$ subset used by the swap analysis. The reference prefix uses the trained boundary markers without latent slots in between. From this reference, we test prefixes that replace the trained markers with two unused special tokens, with the model's $\langle\texttt{bos}\rangle$/$\langle\texttt{eos}\rangle$ pair, with a textual delimiter such as [LATENT]/[/LATENT], with $K$ period or newline tokens that act as pause-like fillers, or with $K$ bare slots and no markers at all. Each marker variant is also tested with and without a length-matched run of latent slots between the markers, which separates the effect of marker identity from the effect of slot-level capacity at fixed length.

\begin{figure*}[t]
\centering
\includegraphics[width=\textwidth]{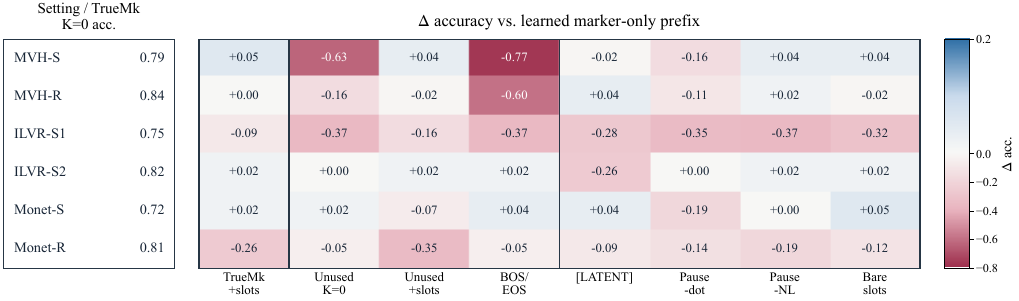}
\caption{\textbf{Dummy-marker forced-prefix controls.} We force a latent-prefix at the end of the prompt and decode from that prefix on V$^*$ ($n=57$). The left column reports accuracy under the learned marker-only prefix; heatmap cells report accuracy differences relative to that condition. Replacing learned markers with unused special tokens or \texttt{<bos>}/\texttt{<eos>} sharply degrades marker-dependent settings, especially \mvh-SFT, showing that learned marker identity is not replaceable by arbitrary special tokens. Adding latent slots or using length-matched bare-slot prefixes can recover some behavior, separating marker-identity routing from latent-slot compute, timing, or formatting capacity.}
\label{fig:app_dummy_marker}
\end{figure*}

Figure~\ref{fig:app_dummy_marker} shows that marker identity matters most when the prefix contains only boundary-like tokens. For \mvh-SFT, replacing learned markers with unused special tokens or \texttt{<bos>}/\texttt{<eos>} causes large drops relative to the true-marker \kzero{} condition. However, adding latent slots can recover much of the behavior even under dummy markers. This distinguishes two mechanisms: a learned routing role tied to marker identity, and a slot-level capacity role provided by latent slots. We therefore avoid the stronger claim that latent slots are useless; the causal claim concerns slot contents as portable visual memories.

\section{Visual-Subspace Alignment and Layer-wise Attention}
\label{app:visual-subspace}

The main text characterizes the relation between slot states and projected visual-token representations along two complementary axes. The first is a static geometric probe. For each V$^*$ sample with emitted latent tokens, we take the final-layer hidden state at every slot position and compare it, in cosine similarity, with every post-merger visual token of the same image after projection into the LLM hidden space. The top-5 mean similarity averages over slots and across the V$^*$ subset (between $14$ and $29$ valid samples per setting depending on the emission rate), giving a single number per setting. ILVR-Stage1 reaches $0.58$, ILVR-Stage2 relaxes to $0.27$, \mvh-SFT and \mvh-RL sit between $0.10$ and $0.15$, and Monet stays near $0.07$, which is indistinguishable from chance-level alignment. The per-setting values for both the coverage and informative halves appear in the concordance table (Appendix~\ref{app:concordance}).

\begin{figure*}[t]
\centering
\includegraphics[width=\textwidth]{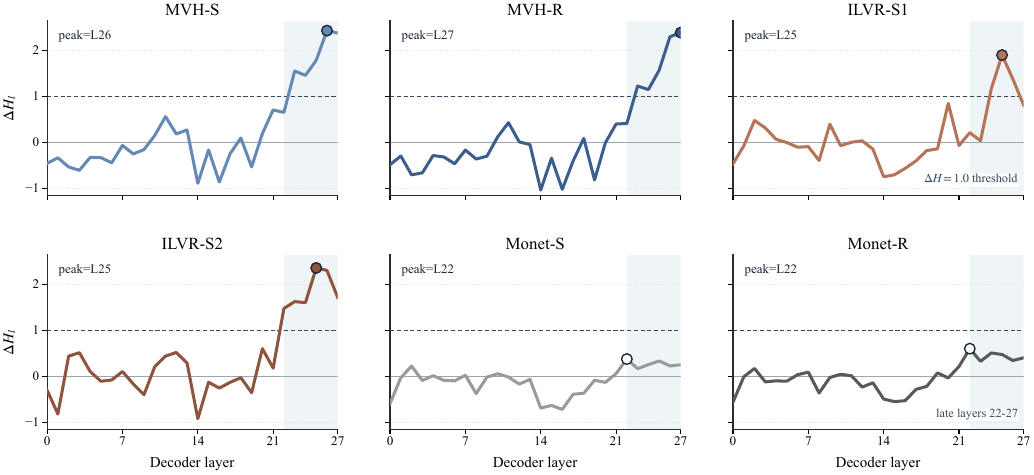}
\caption{\textbf{Full layer-wise attention-concentration curves.} For each method-stage setting, we plot the per-layer image-attention entropy gap $\Delta H_l = H_l^{\mathrm{answer}} - H_l^{\mathrm{latent}}$, where larger values indicate that latent-step attention is more concentrated on image tokens than answer-step attention. The dashed horizontal line marks $\Delta H=1.0$. \mvh-SFT, \mvh-RL, ILVR-Stage1, and ILVR-Stage2 show localized late-layer peaks, with peak layers in L25--L27 and peak gaps between 1.90 and 2.43. Monet-SFT and Monet-RL stay below $\Delta H=1.0$. This expands the compact summary in Figure~\ref{fig:visual-subspace}.}
\label{fig:app_full_layerwise_msl}
\end{figure*}

The second probe is dynamic and decomposes attention to image patches at every decoder layer. For each decoder layer $l$, we compute the entropy gap
\begin{equation}
\Delta H_l = H_l^{\mathrm{answer}} - H_l^{\mathrm{latent}},
\label{eq:delta-h}
\end{equation}
where $H_l^{\mathrm{latent}}$ is the per-layer entropy of decoder attention to image tokens averaged over latent slot positions, and $H_l^{\mathrm{answer}}$ is the analogous quantity averaged over answer-token positions. Low entropy means attention concentrates on a small subset of image patches; positive $\Delta H_l$ means latent-step image attention is more focused than answer-step image attention at layer $l$. The main text reports the peak gap and peak layer for each setting; here we show the full layer-wise curve (Figure~\ref{fig:app_full_layerwise_msl}).

The full curves confirm that the visual-inspection mode is localized rather than uniformly distributed across layers. \mvh{} and ILVR exhibit late-layer peaks, whereas Monet remains below $\Delta H=1.0$. This supports the main-text claim that weak slot-content causality does not imply functional irrelevance: some latent tokens induce a focused visual-inspection mode even when their slot contents do not transfer as recoverable visual memory. All layer-wise attention analyses use the V$^*$ subset used for attention replay; full-sequence HRBench-8K replay was omitted due to memory cost (per-setting values in Table~\ref{tab:concordance-full}).

\subsection{Causal Intervention on Answer-Time Attention}
\label{app:attention-intervention}

The geometry and entropy probes above are diagnostic. To test whether
answer-time visual attention is causally relevant, we start from
\nolatent{}, removing the slots, markers, and latent format, and
intervene only on the answer-token attention distribution at the peak
$\Delta H$ layers. The test uses the V$^*$ latent-gained subset for the
three settings with a clear attention signature. The total attention
mass assigned to image tokens is held fixed, while its distribution is
set to one of three conditions: the model's latent-time distribution
(transplant), an entropy-matched distribution over random locations
(sharp-random), or a distribution over the target patches (oracle).

\begin{table}[t]
\centering
\footnotesize
\setlength{\tabcolsep}{2pt}
\resizebox{\columnwidth}{!}{\input{tables/tabA19_attention_intervention}}
\caption{\textbf{Recovery of latent-gained examples under answer-time
attention intervention (\%).} Brackets are 95\% sample-level bootstrap
confidence intervals ($B=1000$). No steering is zero by construction:
the subset contains examples that \nolatent{} answers incorrectly.}
\label{tab:attention-intervention}
\end{table}

Table~\ref{tab:attention-intervention} shows that transplant recovers
$36\text{--}47\%$ of the latent-gained examples, with every interval
above zero. Changing only the answer-time image-attention distribution
is therefore causally effective. However, transplant and sharp-random
are evaluated on the same examples and differ by only $3\text{--}6$
points (Table~\ref{tab:attention-paired}).

\begin{table}[t]
\centering
\footnotesize
\setlength{\tabcolsep}{3pt}
\resizebox{\columnwidth}{!}{\input{tables/tabA20_attention_paired}}
\caption{\textbf{Paired comparison of transplant and sharp-random
attention.} Every 95\% paired bootstrap interval ($B=1000$) includes
zero.}
\label{tab:attention-paired}
\end{table}

The paired analysis detects no reliable advantage for the model's own
latent-time localization over an equally concentrated random-location
control. This does not establish equivalence or rule out localization
information at the available sample sizes. Oracle attention also
recovers only $49\text{--}60\%$ of the latent-gained sets, so attention
is not an exhaustive account of the gain.

\section{Per-Setting Concordance Across Diagnostics}
\label{app:concordance}

Table~\ref{tab:concordance-full} expands the compact regime summary in Table~\ref{tab:regime-summary} with the underlying per-setting values across all eleven diagnostics used to assign regime labels. The rows group into intervention-based diagnostics (marker sensitivity, malformed interference, marker-only coverage, marker-identity substitution under forced-prefix decoding, swap deviation), generation-behavior diagnostics under marker corruption (self-loop rate, latent blocks emitted), visual-subspace diagnostics (top-5 latent-to-visual similarity for the coverage and informative halves), and attention diagnostics (latent-step image-attention entropy, peak entropy gap). The bottom row reports the final V$^*$ and benchmark-averaged accuracy of each setting. Each column corresponds to one method-stage setting, and the columns are ordered by the marker-dependence ranges defined in \S\ref{sec:regimes}.

\begin{table*}[t]
\centering
\small
\setlength{\tabcolsep}{3pt}
\resizebox{\textwidth}{!}{\input{tables/tabA21_concordance}}
\caption{Full per-setting concordance across independent diagnostics. Slot-content tests, marker interventions, generation behavior, visual-subspace alignment, and attention diagnostics separate the six method-stage settings into three marker-dependence ranges. The bottom row shows that final accuracies are similar; the other rows show that the settings use the latent interface differently. Compact summary appears in Table~\ref{tab:regime-summary}.}
\label{tab:concordance-full}
\end{table*}

The bottom-row accuracies are similar across columns, while the upper rows separate the three ranges along multiple axes. This is the form of evidence that supports the two-axis description in \S\ref{sec:regimes}: final accuracy alone does not identify how a given latent-token setting is being used.

\section{AI Assistance Statement}

We used ChatGPT solely for light language polishing (grammar, word
choice, and phrasing) on portions of the manuscript. All research
ideas, experimental design, implementation, analysis, and scientific
claims are the authors' own. The authors verified all AI-assisted
text edits for accuracy and take full responsibility for the
content of the paper.

\end{document}

%% file: 0_abstract.tex
\begin{abstract}
Recent latent visual reasoning methods achieve substantial gains
by inserting continuous latent tokens into multimodal language
models. These gains are commonly attributed to the tokens encoding
visual evidence; recent analyses, however, reveal a paradox: the
tokens are loosely tied to the image and contribute little to the
answer. Critically, these analyses treat latent tokens as a single
unit, obscuring the source of the gains. We therefore
decompose latent tokens into three testable components: latent
slots, boundary markers, and format, and develop a state-of-the-art
method as a probe under favorable conditions. Across six
method-stage settings and four perception-heavy benchmarks, latent
slots fail every prediction of the visual-memory account.
Strikingly, retaining only the boundary markers preserves 78 to
100\% of the gain in several settings, while the model attends to
the image more narrowly at latent positions than at answer
positions. These results do not support slot contents as recoverable visual memory; much of the benefit is instead associated with marker and format control, together with visual-attention routing. At matched accuracy, methods can still rely on markedly different mechanisms shaped by training supervision. Latent visual reasoning thus needs evaluation not only by accuracy but by what the model actually relies on.
\end{abstract}

%% file: 1_introduction.tex
\section{Introduction}

\begin{figure}[t]
\centering
\includegraphics[width=\columnwidth]{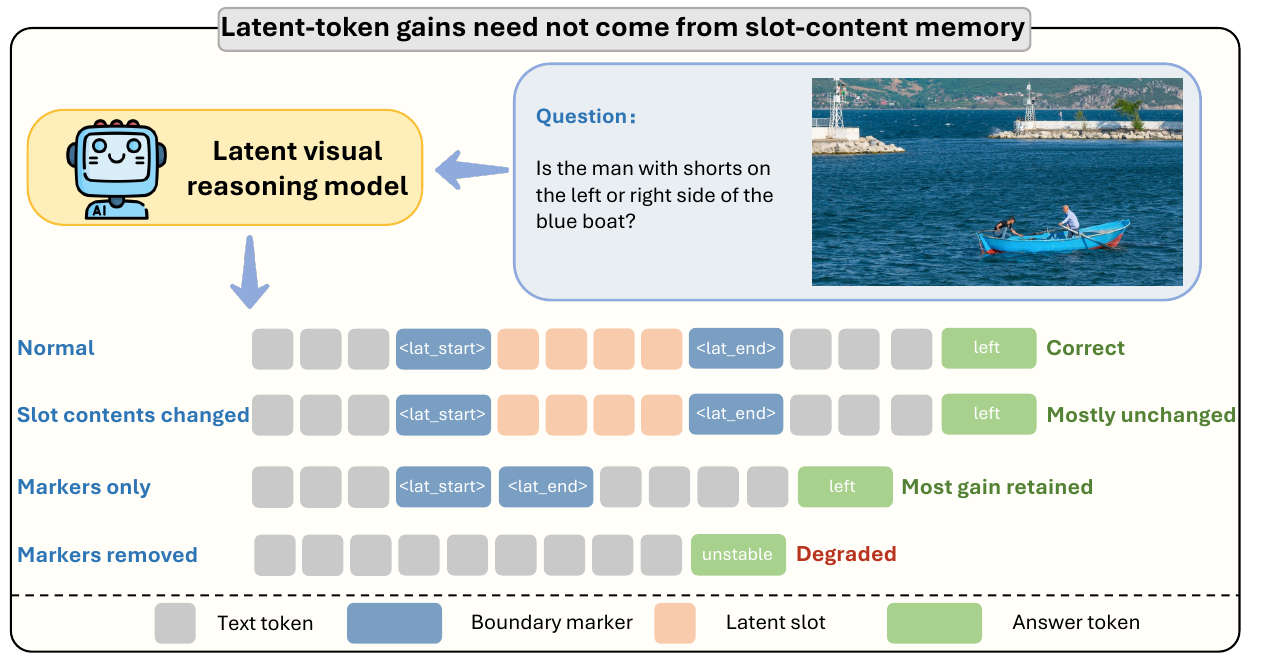}
\caption{
\textbf{Latent-token gains need not come from slot-content memory.}
Changing slot contents has little effect, while keeping only the
learned boundary markers preserves $78\text{--}100\%$ of the
latent-token gain in most \mvh{} and ILVR settings. Corrupting the
boundary markers can instead cause degenerate generation, suggesting
that latent-token gains can arise from boundary markers and format
rather than from recoverable slot contents.
}
\label{fig:teaser}
\end{figure}

Visual reasoning often relies on signals that resist textual encoding: 
fine-grained spatial relations, localized evidence, and intermediate 
visual hypotheses lose their structure once verbalized into a 
textual chain of thought \citep{NEURIPS2024_fb820110, Man_2025_CVPR, NEURIPS2022_9d560961}. 
Recent methods therefore insert latent 
tokens into multimodal language models, conditioning final answers 
on intermediate hidden states rather than text alone \citep{Bigverdi_2025_CVPR}, with a common 
design wrapping a span of latent tokens with two learned delimiters:
\begin{equation}
\langle\texttt{lat\_start}\rangle, s_1,\ldots,s_K,
\langle\texttt{lat\_end}\rangle .
\label{eq:latent-tokens}
\end{equation}
The learned delimiters are defined as \emph{boundary markers}, while the 
intervals between them are defined as \emph{latent slots}.
The latent slots 
condition the following answer and are commonly assumed to encode 
visual evidence beyond text, working like an internal sketchpad. 
We term this theoretical framework the \emph{visual-memory account}.

Recent analyses challenge this account, finding latent tokens only
loosely tied to the image and only weakly causal for the
answer~\citep{zhang2026visuallatentsknowsay, li2026imagination,
zhang2025latenttokensthinkcausal}. Yet they treat latent tokens as
a single unit, leaving open which component is at fault and whether
any remains functional. Final accuracy cannot decide: the same gain
could come from latent slots, boundary markers, format, or
attention shifts during latent generation.

We therefore decompose latent tokens into three independently
testable components: latent slots, boundary markers, and format,
and intervene on each in isolation. To test the visual-memory
account under conditions favorable to it, we develop \textbf{\mvh{}}
(Mesoscale Visual Hypotheses), a \emph{content-favorable probe}
that supervises slot contents with broad image coverage and
question-relevant anchors. Under matched data, backbone, and
hyperparameters, \mvh{} reaches state-of-the-art accuracy among the
latent-visual methods we compare
(Table~\ref{tab:accuracy-setup}): if slot contents ever encode
recoverable visual evidence, they should encode it here.

Across six method-stage settings and four perception-heavy
benchmarks, slot contents fail every prediction of the
visual-memory account: changing them moves accuracy by at most
$0.027$, image-conditioned slot states recover only $7.8\%$ of the
missing-image gap, and pooled cross-example swaps transfer donor answers
at $5.3\text{--}12.7\%$, below chance.
Strikingly, retaining only
the boundary markers preserves $78\text{--}100\%$ of the
latent-token gain in several settings, while at the same positions
the model attends to the image more narrowly than during answer
generation. Together, these results do not support recoverable
slot-content memory; much of the benefit is instead associated with
marker and format control and focused visual attention during latent generation
(Figure~\ref{fig:teaser}).

Settings with near-identical accuracy further separate along two
diagnostic axes: dependence on boundary markers and format, and
proximity between slot contents and projected visual-token
representations. Methods that directly anchor slots to visual-token
representations develop strong marker dependence, while methods
that distill from teacher hidden states do not. Two methods can
match in accuracy yet rely on latent tokens through markedly
different mechanisms.

\paragraph{Contributions.}
\begin{enumerate}[leftmargin=*,topsep=2pt,itemsep=2pt]
    \item We decompose latent tokens into three independently
    testable components: latent slots, boundary markers, and
    format, moving the analysis from a single-unit view to a
    component-level one.
    \item We develop \mvh{}, a content-favorable probe that reaches
    state-of-the-art accuracy under matched settings, and use it to
    give the visual-memory account its strongest test.
    \item Beyond the negative slot-memory result, we identify major
    non-content mechanisms associated with the gain: marker and format control, together with visual-attention routing.
    Their use varies with training supervision, motivating
    mechanism-level evaluation beyond accuracy.
\end{enumerate}

%% file: 2_related.tex
\section{Related Work}
\label{sec:related}

\begin{figure*}[t]
    \centering
    \includegraphics[width=\textwidth]{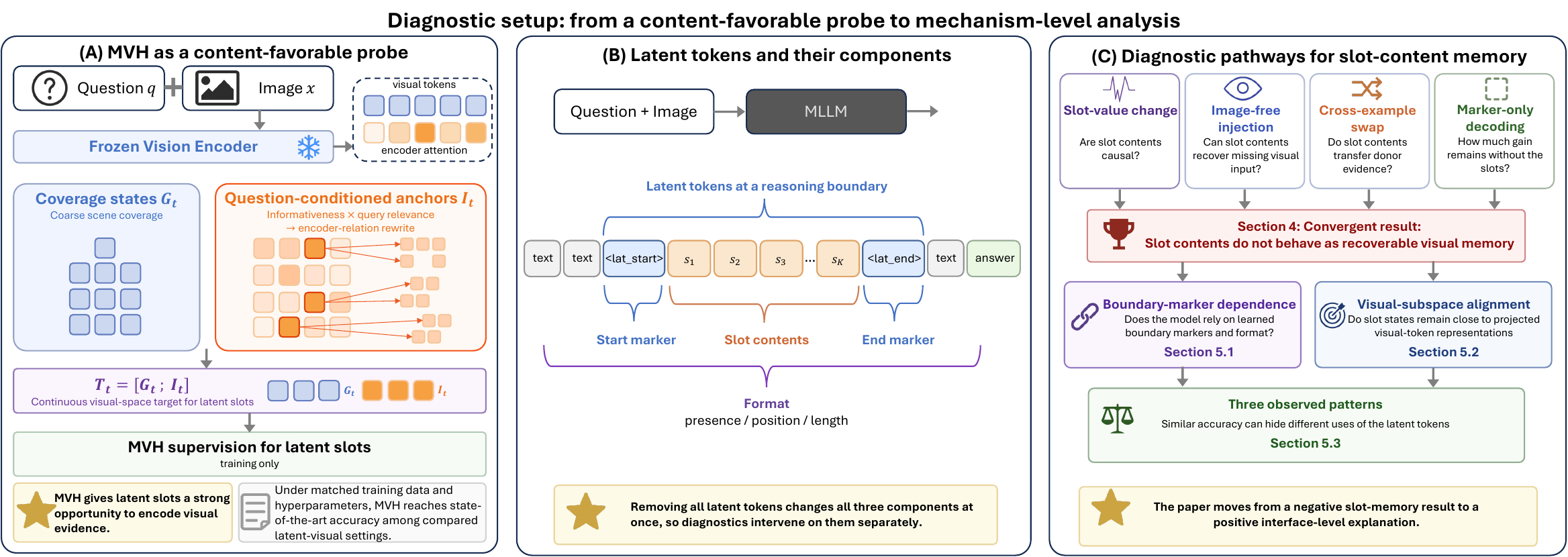}
    \caption{
    \textbf{Diagnostic setup.}
    (A) \mvh{} trains latent slots with visual targets that combine broad
    image coverage and question-relevant anchors.
    (B) Latent tokens have three components that can be intervened on
    separately: slot contents, boundary markers, and format.
    (C) We first test whether slot contents behave as recoverable visual
    memory, then examine how the latent interface is used through boundary-marker
    dependence and visual-subspace alignment.
    }
    \label{fig:diagnostic-setup}
\end{figure*}

\paragraph{Latent visual reasoning methods.}
Recent methods insert continuous visual states into multimodal LLM
decoding. Mirage grounds latent tokens via image-embedding supervision
and aligns the trajectory with task
objectives~\citep{yang2026machine}; LVR interleaves latent and text
generation to reconstruct query-relevant visual
tokens~\citep{li2026latent}; ILVR adds selective perceptual modeling
under a teacher
model~\citep{dong2026interleavedlatentvisualreasoning}; and Monet uses
teacher-state distillation refined by a visual-latent policy
objective~\citep{wang2026monet}. Other methods enrich the supervision
signal in different
ways~\citep{qin2025chainofvisualthoughtteachingvlmsthink,chen2026think,wang2026foresttreeslatentsuperposition,ding2026colvrenhancingexploratorylatent,NEURIPS2025_95c6ae3f},
while a separate line externalizes visual reasoning through sketches,
pixel-level renderings, or tool-assisted
inspection~\citep{zhang2025latentsketchpadsketchingvisual,li2025imagine,zheng2026deepeyes}.
Across continuous-latent designs, slots are typically described as
carrying task-relevant visual evidence; we test whether these hidden
states actually behave as recoverable visual memory rather than propose
yet another such design.

\paragraph{Analyses and critiques of latent visual reasoning.}
Recent analyses challenge this picture.
CapImagine~\citep{li2026imagination} reports that inputs have little
effect on latent states and that latent states have little effect on
the answer, and shows explicit text-based imagination matches or
exceeds latent-space methods on vision-centric benchmarks.
\citet{viveiros2026whatsholdinglatentvisual} find that uninformative
dummy tokens leave accuracy nearly unchanged, tracing this to oracle
latents carrying little information and inference-time latents
collapsing to a narrow region. \citet{zhang2026visuallatentsknowsay}
identify a ``silenced visual latents'' phenomenon and propose
inference-time optimization to recover the suppressed contributions.
Our work differs in two ways: we build \mvh{}, which reaches
state-of-the-art accuracy under matched settings with structured visual
targets, so the visual-memory account is tested under favorable
conditions; and by intervening on slot contents, boundary markers, and
format separately, paired with visual-subspace and image-attention
probes, we move from diagnosing failure to characterizing how latent
tokens are used when slot-content memory fails. This is complementary
to CapImagine: high-accuracy latent methods do not necessarily achieve
their gains through the content storage their design suggests.

\paragraph{Latent reasoning and structural tokens.}
Our results connect to broader work on tokens that carry no textual
content~\citep{zhu2025surveylatentreasoning}: continuous
thoughts~\citep{hao2025traininglargelanguagemodels}, pause
tokens~\citep{goyal2024think}, filler tokens~\citep{pfau2024lets}, and
register tokens in vision transformers~\citep{darcet2024vision} all
show that non-content tokens can shape behavior. Latent tokens in a
multimodal model, however, combine slot contents, boundary markers,
format, and a geometry tied to the visual encoder. The decomposition
into these components, and the separation of marker-driven control
from slot-driven content, have no direct analogue in the text-only
setting; we examine each component in Section~\ref{sec:pad-tests}.

%% file: 3_method.tex
\section{Diagnostic Setup}
\label{sec:setup}

We first describe \mvh{} as a favorable probe
(\S\ref{sec:mvh-probe}), then specify the settings, benchmarks, and
tests used in the analysis (\S\ref{sec:settings-diagnostics}).
Figure~\ref{fig:diagnostic-setup} gives an overview.

\subsection{MVH: A Content-Favorable Probe}
\label{sec:mvh-probe}

Earlier latent-visual methods supervise slot states at different levels
of detail. Patch-level alignment ties each slot to a specific visual
region, while distillation from teacher states provides a flexible but
less explicit visual signal. \mvh{} sits between these two extremes:
each slot is trained toward a target that combines broad image coverage
with question-relevant anchors (Figure~\ref{fig:diagnostic-setup}A).
The design intentionally gives the visual-memory account a structured,
explicitly visual target. If slot contents can ever encode recoverable
visual evidence, \mvh{} should be one of the easier settings in which
to observe it.

\paragraph{Target structure.}
At each reasoning step $t$, the frozen vision encoder produces visual
tokens $V_t = \{v_{t,1}, \ldots, v_{t,M_t}\}$ on an $H_t \times W_t$
grid, with self-attention maps $\{A_t^{(h)}\}$. The teacher target is
\begin{equation}
  T_t = [G_t;\, I_t], \qquad |T_t| = K_g + K_i = N,
\label{eq:teacher-target}
\end{equation}
where $G_t$ contains $K_g$ coverage states (the global mean of $V_t$
plus nine regional means over a $3\times 3$ partition) and $I_t$
contains $K_i$ question-relevant anchors. We use
$K_g = K_i = 10$ and latent tokens of length $N = 20$ at training time.

\paragraph{Selecting anchors.}
Anchors are picked by a fused score
\begin{equation}
  p_{t,j} \propto r_j^{\text{base}} \cdot u_{t,j},
\label{eq:fused-score}
\end{equation}
where $r_j^{\text{base}}$ measures the average attention each visual
token receives from the encoder, and $u_{t,j}$ measures the relevance
of that token to the current decoder context. The top-$K_i$ tokens are
then rewritten into compositional anchors that mix each selected token
with its encoder-attention neighbors, so anchors reflect contextual
co-evidence rather than purely appearance-based redundancy. Full
equations for the score, the rewriting kernel, the EMA teacher, and
norm-aligned injection are in Appendix~\ref{sec:method-appendix}.

\paragraph{SFT stage.}
The SFT stage supervises each latent slot toward its target via
next-token-shifted cosine alignment:
\begin{equation}
\begin{split}
  \mathcal{L}_{\text{lat}}
    = \frac{1}{|\mathcal{S}_{\text{lat}}|} \!\!
      \sum_{(t,i)\in\mathcal{S}_{\text{lat}}} \!\!
      \bigl(1 - \cos(z_{t,i-1},\, \tau_{t,i}^{\text{aligned}})\bigr),
\end{split}
\label{eq:sft-latent}
\end{equation}
where $z_{t,i-1}$ is the student's last-layer hidden state at the
position immediately before the $i$-th slot, and
$\tau_{t,i}^{\text{aligned}}$ is the norm-rescaled teacher target.
Cosine alignment is used because norm-aligned injection already matches
the scale of the teacher to the embedding space, leaving direction as
the only degree of freedom. The full SFT objective combines this loss
with the standard text cross-entropy under a linear warm-up.

\begin{table}[t]
\centering
\footnotesize
\setlength{\tabcolsep}{2pt}
\input{tables/tab1_setup_accuracy}
\caption{
\textbf{Accuracy of the analyzed settings.}
SFT-stage settings share the same backbone, SFT data, and hyperparameters.
For RL-stage settings, Monet's RL data is not public; we train
\mvh{}-RL using data built following Monet's construction pipeline.
\mvh{} reaches the strongest accuracy among the latent-visual settings
we analyze, supporting its use as a favorable probe. The \textbf{bold numbers} indicate the best performance achieved by each setting.}
\label{tab:accuracy-setup}
\end{table}

\paragraph{RL stage.}
The RL stage refines the policy with an outcome reward while preserving
the slot geometry learned during SFT. We sample $G$ rollouts per prompt
and compute GRPO~\citep{shao2024deepseekmathpushinglimitsmathematical} advantages $\hat A_g$ from a reward mixing answer
correctness and format. Text tokens follow the standard clipped PPO
objective~\citep{schulman2017proximal}. Slot states are continuous and have no token-level
log-probability, so instead of fitting a parametric density over them
as in VLPO~\citep{wang2026monet}, we propagate each rollout's advantage through
cosine alignment between the on-policy slot state $z^\theta_{t,i}$ and
the rollout-time state $z^{(g)}_{t,i}$:
\begin{equation}
  \mathcal{L}_{\text{lat}}^{\text{RL}}
    = -\,\mathbb{E}_{g,(t,i)}\!
      \bigl[\hat A_g \cdot \cos(z^\theta_{t,i},\, z^{(g)}_{t,i})\bigr].
\label{eq:rl-latent}
\end{equation}
A latent trust region anchored to the frozen SFT checkpoint, the
continuous analogue of PPO's reference-KL, prevents drift from the SFT
subspace. Full objective and weights are in
Appendix~\ref{sec:method-appendix}.

\paragraph{Inference.}
No teacher is needed at inference: the model generates the latent tokens
autoregressively from image and question. We use $K_{\text{eval}} = 8$
latent slots per step in the primary evaluations.

\paragraph{Why \mvh{} is a favorable probe.}
Two pieces of evidence support this role. First, \mvh{}-RL reaches
state-of-the-art accuracy among the latent-visual settings we compare
(Table~\ref{tab:accuracy-setup}).
Second, the latent slots produced by
\mvh{} respond to the input: their cross-sample cosine similarity
averages $0.68$, well below the $0.96\text{--}0.97$ observed for ILVR
and Monet (Figure~\ref{fig:app_structural_diagnostics}A), showing that
they vary with the input rather than collapsing across examples. Within
a sample, coverage and anchor slots separate into distinct sub-blocks
($\mathrm{BSR}=1.13$), and on the same image with different questions,
the anchor slots shift more than the coverage slots
($\mathrm{RSR}=1.12$, against $\approx 1.00$ for baselines). Accuracy
alone would be a circular justification for the probe; the structural
diagnostics above are independent of accuracy, and they show that
\mvh{}'s slots respond to the input and to question variation in ways
the visual-memory account would predict.

\begin{figure*}[t]
\centering
\includegraphics[width=\textwidth]{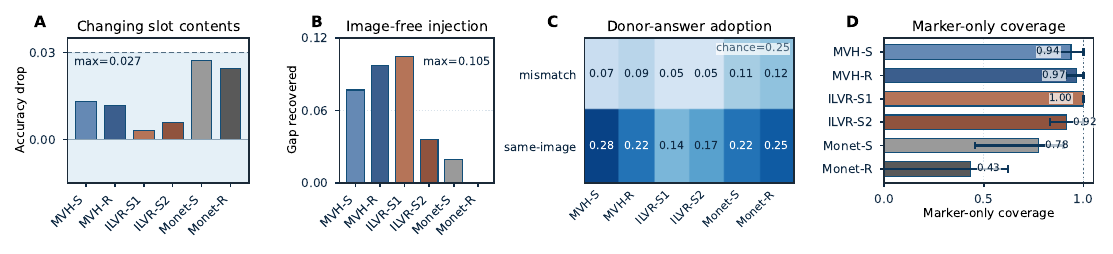}
\caption{\textbf{Four tests of slot-content memory.}
(A) Changing slot contents (zero, random, or fixed substitutes) produces
only small accuracy changes. (B) Injecting image-conditioned slot
states into image-free runs recovers little of the missing-image gap.
(C) Swapping real slot states across examples does not reliably
transfer donor-specific answers; donor-answer adoption stays near or
below chance, including in same-image cross-question controls. Pooled
four-benchmark results and bootstrap intervals are reported in
Appendix~\ref{app:slot-swap}. (D)
Marker-only decoding recovers much of the unique latent-token gain over
no-latent decoding. Small effects in (A)--(B), donor adoption near
chance in (C), and high marker-only coverage in (D) do not support the
view that slot contents behave as recoverable visual memory, even under
content-favorable \mvh{} supervision.}
\label{fig:causal-tests}
\end{figure*}

\subsection{Settings, Benchmarks, and Tests}
\label{sec:settings-diagnostics}

Latent tokens (Figure~\ref{fig:diagnostic-setup}B) have three components
that can be intervened on separately: the slot contents, the boundary
markers, and the latent format (its presence, position, and length). We
exploit this separability throughout the analysis. Each test in
Sections~\ref{sec:pad-tests}--\ref{sec:mechanism-space} changes one
component at a time and looks at how accuracy or generation behavior
responds.

\paragraph{Settings.}
We analyze six method-stage settings across three supervision
strategies. \mvh{}-SFT and \mvh{}-RL correspond to the two stages
of \mvh{} described above. The other two methods are chosen for
their representativeness: ILVR adopts direct visual-token alignment
and Monet adopts teacher-state distillation, the two dominant
supervision strategies in latent visual reasoning, and both achieve
strong performance among published methods. ILVR-Stage1 and
ILVR-Stage2 differ in that Stage2 relaxes the alignment target;
Monet-SFT and Monet-RL differ in that the RL stage applies
reward-based optimization on answer correctness. All six settings
share the same latent-token structure but differ in how the slot
states are supervised.

\paragraph{Training data.}
Monet checkpoints come from the original authors. We re-implement and
train the other four settings; \mvh{}-SFT, ILVR-Stage1, and ILVR-Stage2
share the SFT corpus released by Monet, making the SFT comparison fully
matched in data. Monet's RL corpus is not released; we follow its
pipeline to build one from DeepEyes-v2 and use it to train \mvh{}-RL.
The original RL comparison is therefore not data-matched, but retraining
Monet on the exact \mvh{}-RL data leaves its performance essentially
unchanged (Appendix~\ref{app:monet-matched}). Reproduction details
appear in Appendix~\ref{app:training-data}.

\paragraph{Benchmarks.}
We evaluate on V$^*$~\citep{wu2024v}, HRBench-4K, HRBench-8K~\citep{wang2025divide}, and MME-RealWorld-Lite~\citep{zhang2025mmerealworld}.
This suite stresses fine-grained recognition, high-resolution
inspection, spatial reasoning, and real-world visual understanding, and
is well suited for testing whether latent states carry task-relevant
visual evidence.

\begin{figure*}[t]
\centering
\includegraphics[width=\textwidth]{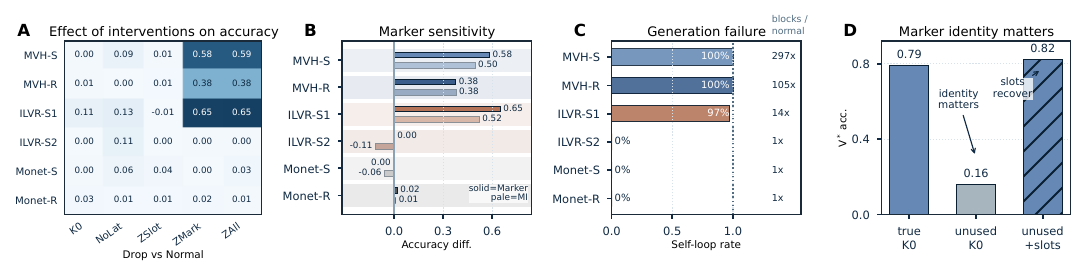}
\caption{
\textbf{Marker-dependence diagnostics.}
(A) Accuracy drops under component interventions. (B) Marker
sensitivity (solid) and malformed-marker interference (pale). (C)
Self-loop rates after marker zeroing, with increases in generated latent
blocks at right. (D) \mvh{}-SFT forced-prefix controls with learned or
unused markers, with and without slots. Larger drops and self-loop rates
indicate stronger marker dependence. Changing slot contents has little
effect, whereas marker zeroing disrupts the marker-dependent settings,
especially \mvh{}-SFT and ILVR-Stage1. K0 keeps markers but removes
slots; NoLat removes all latent tokens; ZSlot zeros slot states; ZMark
zeros marker embeddings; ZAll zeros both. Full condition definitions
appear in Table~\ref{tab:app_intervention_conditions}.
}
\label{fig:marker-protocol}
\end{figure*}

%% file: tables/tab1_setup_accuracy.tex
\begin{tabular}{@{}lccccc@{}}
\toprule
\textbf{Setting} & \textbf{V$^*$} & \textbf{HR-4K} & \textbf{HR-8K} & \textbf{MME-RW} & \textbf{Avg.} \\
\midrule
\rowcolor[HTML]{E1E1E1}
\multicolumn{6}{c}{\textit{\textbf{Proprietary Model}}} \\
GPT-4o                   & 67.5 & 59.0 & 55.5 & 52.0 & 58.5 \\
\midrule
\rowcolor[HTML]{E1E1E1}
\multicolumn{6}{c}{\textit{\textbf{Open-Source Baselines}}} \\
Qwen2.5-VL-7B          & 76.4 & 68.0 & 63.8 & 45.8 & 63.5 \\
\quad + vanilla SFT    & 81.7 & 68.4 & 61.6 & 51.3 & 65.8 \\
\quad + SFT + GRPO     & 78.5 & 70.0 & 66.8 & 52.4 & 66.9 \\
ILVR-Stage1            & 80.6 & 70.0 & 66.5 & 49.2 & 66.6 \\
ILVR-Stage2            & 80.6 & 70.3 & 66.6 & 50.4 & 67.0 \\
Monet-SFT              & 79.6 & 70.4 & 63.0 & 52.9 & 66.5 \\
Monet-RL               & 82.2 & 71.0 & 66.0 & 51.2 & 67.6 \\
\midrule
\rowcolor{ourblue}
\multicolumn{6}{c}{\textit{\textbf{Our Model}}} \\
\textbf{\mvh-SFT}      & 82.2 & 71.5 & 67.1 & 50.4 & 67.8 \\
\textbf{\mvh-RL}       & \textbf{83.2} & \textbf{72.4} & \textbf{69.5} & \textbf{53.8} & \textbf{69.7} \\
\bottomrule
\end{tabular}

%% file: 4_results.tex
\section{Slot Contents Do Not Behave as Recoverable Visual Memory}
\label{sec:pad-tests}

We now test the visual-memory account on the probe established in
Section~\ref{sec:setup}. If slot contents act as recoverable visual memory, they should be 
causal for the answer, recoverable when the image is removed, 
portable across samples, and responsible for the unique latent-token 
gain. We
test these four predictions with four complementary tests
(Figure~\ref{fig:causal-tests}).

\subsection{Four Tests of Slot-Content Memory}
\label{sec:four-tests}

\paragraph{Changing slot contents.}
The first test replaces the slot contents with
substitutes (zero, random, or fixed vectors) while keeping the
boundary markers in place. If slot contents carry the evidence used by
the decoder, these substitutions should hurt accuracy. They do not.
Benchmark-averaged accuracy moves by at most $0.027$ across the six
settings (Figure~\ref{fig:causal-tests}A), even though the same models
respond strongly to other interventions on the latent tokens
(Section~\ref{sec:mechanism-space}). Slot contents are weakly causal
under direct perturbation.

\paragraph{Injecting latents into image-free runs.}
A natural follow-up asks whether slot contents can recover visual
information when the image itself is removed. We run this test across
the method-stage settings in Figure~\ref{fig:causal-tests}B. For each
example, we blank the image and inject the slot states that the same
model produced with the image present. If latent
tokens carry recoverable visual content, these states should close a
substantial fraction of the gap between blind and full-image decoding.
Across settings, injected image-conditioned states recover only a small
fraction of the missing-image gap. For example, \mvh{}-SFT reaches
$0.845$ with the image and $0.228$ without it; injected states raise
blind accuracy only to $0.276$, recovering $7.8\%$ of the gap
(Figure~\ref{fig:causal-tests}B). Its estimate is imprecise because the
valid subset is small, so we treat it as supporting rather than decisive evidence
(Appendix~\ref{app:blind-injection}).

\paragraph{Swapping slot states across examples.}
Zero, random, and fixed substitutes are out-of-distribution. This test
removes that concern: slot states are captured from real model runs on
donor examples and injected at the corresponding decoder layer of a
target run, with the layer swept across three depths
(Appendix~\ref{app:slot-swap}). If slot states behave as
recoverable visual memory, a real donor state should transfer some of
the donor's content, especially when donor and target share the same
image but ask different questions. This same-image cross-question
control is favorable to the visual-memory account: donor and target
share the visual scene and differ mainly in the question-dependent
evidence required to answer. Pooled across all four benchmarks,
donor-answer adoption remains $5.3\text{--}12.7\%$, with every 95\%
bootstrap interval below the $25\%$ chance rate
(Appendix~\ref{app:slot-swap}). Slot states transfer little
donor-specific information.

\paragraph{Marker-only decoding.}
The final test moves from aggregate accuracy to the examples on which
latent tokens are most clearly useful: those answered correctly by
normal latent decoding, but missed when we drop the latent tokens
entirely. We ask, of these examples, how many remain solved if we
keep only the boundary markers and drop the slots between them. If
slot contents carry the unique latent-token contribution, this fraction
should be low. Instead, marker-only decoding recovers $78\text{--}100\%$
of these gained examples in most \mvh{} and ILVR
settings~(Figure~\ref{fig:causal-tests}D).\footnote{Monet-RL is an
exception; per-setting values and small-denominator cases are in
Appendix~\ref{app:moc}.} The unique latent-token contribution is
therefore largely recoverable without the slot states that sit between
the markers.

\subsection{Evidence Across Tests}
\label{sec:convergent}

The four tests make different assumptions but point in the same direction
(Figure~\ref{fig:causal-tests}): perturbation tests causal sensitivity,
image-free injection tests recoverability, swaps test portability, and marker-only decoding tests responsibility for the unique latent-token gain.
Together, slot contents show little perturbation sensitivity, do not substantially recover missing visual input, show little donor-specific transfer across examples, and are not needed for most of the unique latent-token gain. These results do not support the visual-memory account, even when latent
slots are trained with content-favorable \mvh{} supervision.

%% file: 5_analysis.tex
\section{How Latent Tokens Are Used Without Slot-Content Memory}
\label{sec:mechanism-space}

Section~\ref{sec:pad-tests} establishes two anchors for what follows:
slot contents do not behave as recoverable visual memory, while marker-only
decoding preserves much of the latent-token gain. Two questions
follow. Are boundary markers interchangeable placeholders or learned
control signals, and does the answer hold across methods? Beyond the
markers, does latent generation itself carry any role? Section~\ref{sec:marker-protocol}
characterizes the markers, Section~\ref{sec:visual-subspace} examines
visual behavior during latent generation, and Section~\ref{sec:regimes}
combines the two axes.

\subsection{Boundary Markers Are Control Signals, Not Placeholders}
\label{sec:marker-protocol}

\noindent\textbf{Boundary markers act as learned control signals
for entering and leaving latent generation, not as interchangeable
placeholders. Whether they take this role depends on training
supervision.} Two observations support this claim, both from
perturbing the markers while leaving slot states and format intact.

\textbf{Observation 1: marker perturbation produces a sharply
polarized effect.} Zeroing marker embeddings causes \mvh{}-SFT and
ILVR-Stage1 to collapse, with benchmark-averaged drops of $0.585$
and $0.651$; \mvh{}-RL loses $0.378$, while ILVR-Stage2 and Monet
remain essentially unaffected ($\leq 0.02$). The same polarization
appears when the trained markers are replaced by unused special
tokens or \texttt{<bos>}/\texttt{<eos>}
(Appendix~\ref{app:marker-controls}), indicating that the trained
markers carry information that arbitrary special tokens do not
reproduce. The same latent format therefore supports markedly
different modes of use across training settings; pure placeholders
would not yield such polarization.

\textbf{Observation 2: zeroing markers induces self-looping
generation.} In marker-dependent settings, zeroing the markers
leads \mvh{}-SFT and ILVR-Stage1 to enter self-looping latent
generation in nearly all V$^*$ runs, repeatedly emitting latent
insertions without terminating. This is not attributable to a
generic loss of computation, as the latent slots remain present,
and is consistent with disrupted control over how the decoder
enters and exits latent mode.

Monet, where markers carry no such role, is not a counterexample
but a prediction of this account: its training never anchors slot
states to visual targets, so the markers are never recruited as
mode-switching signals. We return to this supervision-mechanism
link in Section~\ref{sec:regimes}. This contrast parallels findings
from \emph{Let's Think Dot by Dot} \citep{pfau2024lets}, which shows that content-free filler
tokens can support hidden computation in text-only Transformers.
Our results extend this picture along two multimodal-specific
dimensions: markers and slot contents play asymmetric roles within
the same token, and latent generation exhibits a visual behavior
without a text-only analogue, which we examine next.

\begin{figure}[t]
\centering
\includegraphics[width=\columnwidth]{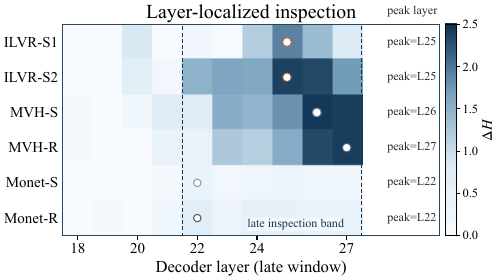}
\caption{
\textbf{Layer-localized visual inspection.}
$\Delta H_l = H_l^{\mathrm{answer}} - H_l^{\mathrm{latent}}$;
positive values mean image attention is more concentrated at latent
than at answer positions. \mvh{} and ILVR show clear late-layer
peaks; Monet does not. This figure is diagnostic; a direct attention
intervention is reported in Appendix~\ref{app:attention-intervention}.
}
\label{fig:visual-subspace}
\end{figure}

\subsection{Visual Engagement Without Recoverable Slot Memory}
\label{sec:visual-subspace}

\noindent\textbf{Latent generation engages the image, but its slot
states do not behave as recoverable visual memory under our tests
(Section~\ref{sec:pad-tests}).} We examine geometry and attention.

\textbf{Probe 1 (geometric): slot states lie close to visual-token
representations in a method-dependent order.} For each slot position
we compute the cosine similarity between its final-layer hidden
state and each projected visual-token representation, retain the
top-5 values, and average across slots. The settings form a clear
order: ILVR-Stage1 remains closest to visual-token representations,
ILVR-Stage2 relaxes this alignment, \mvh{} occupies a more
compressed region of visual space, and Monet sits near the
low-similarity end. A slot state can therefore be visually aligned
without functioning as portable memory.

\textbf{Probe 2 (dynamic): the model attends to the image more
narrowly during latent generation than during answer generation.}
At decoder layer $l$, $H_l^{\mathrm{latent}}$ and
$H_l^{\mathrm{answer}}$ measure the entropy of decoder attention to
image tokens, averaged over slot positions and answer-token
positions respectively (Appendix~\ref{app:visual-subspace}). A
positive $\Delta H_l = H_l^{\mathrm{answer}} - H_l^{\mathrm{latent}}$
indicates attention is more concentrated during latent generation.
\mvh{} and ILVR exhibit clear late-layer peaks, with peak $\Delta H$
ranging from $1.90$ to $2.43$ between decoder layers $25$ and $27$
(Figure~\ref{fig:visual-subspace}). Monet's nominal peak appears at
layer $22$ but remains below $\Delta H = 1.0$.

Transplanted latent-time attention is causally effective, but not reliably
better than an equally concentrated random-location control
(Appendix~\ref{app:attention-intervention}).

Overall, \mvh{} and ILVR show a layer-localized visual-inspection signature,
while Monet does not; the resulting states still do not behave as portable
visual memory.

\begin{figure}[t]
\centering
\includegraphics[width=\columnwidth]{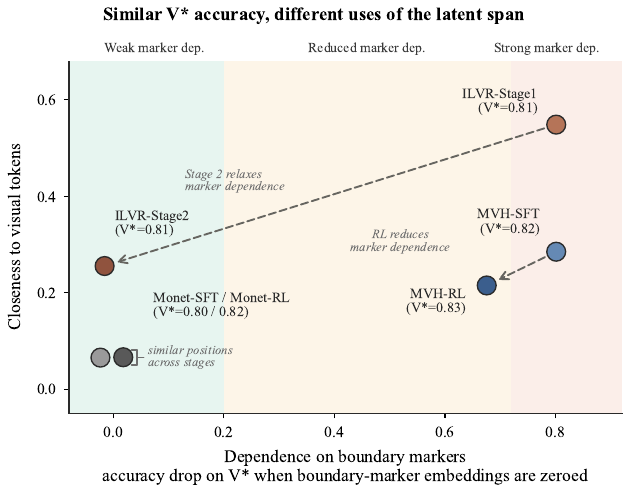}
\caption{
\textbf{Similar V$^*$ accuracy, different uses of latent tokens.}
The six settings reach similar V$^*$ accuracy but separate along
boundary-marker dependence and closeness to visual-token
representations. Moving right means a larger V$^*$ accuracy drop when
boundary-marker embeddings are zeroed; moving up means greater top-5
similarity to visual-token representations. Shaded regions are coarse
descriptive ranges, not method categories. Arrows show how later
training stages move settings along these axes.
}
\label{fig:mechanism-space}
\end{figure}

\subsection{Same Accuracy, Different Mechanisms}
\label{sec:regimes}

\noindent\textbf{At matched accuracy, methods rely on latent tokens
through markedly different mechanisms, and these mechanisms are
shaped by training supervision.} The two diagnostic axes vary
sharply across the six settings: marker drop ranges from $\leq 0.02$
to $0.651$, and top-5 visual similarity from below $0.10$ to above
$0.50$, even though V$^*$ accuracy remains within
$0.80\text{--}0.83$ (Figure~\ref{fig:mechanism-space},
Table~\ref{tab:regime-summary}).

\textbf{Finding 1: training supervision places methods on both axes.}
\mvh{}-SFT and ILVR-Stage1 anchor slot states directly to
visual-token representations. The model learns to switch into a
visual processing mode at the markers, producing strong marker
dependence ($0.585$ and $0.651$); slot states are pulled into the
visual subspace, so latent generation is accompanied by focused
image attention ($\Delta H \geq 1.90$). Monet distills teacher
hidden states instead, so markers are never recruited as control
signals and latent generation shows no comparable focused-attention
signature; both axes remain low (marker dependence $\leq 0.02$,
$\Delta H < 1.0$). RL acts as a relaxer rather than a builder: in \mvh{}-RL,
refinement on top of \mvh{}-SFT reduces marker dependence from
$0.585$ to $0.378$ without driving it to zero.

\textbf{Finding 2: the two axes can vary independently.} The
clearest evidence comes from ILVR's Stage1-to-Stage2 transition:
relaxing the visual-alignment target collapses marker dependence
from $0.651$ to $\leq 0.02$, yet the late-layer visual-inspection
signature is preserved ($\Delta H \geq 1.90$). A setting can
therefore lose its reliance on markers while retaining focused
image attention.

Two methods can thus reach near-identical accuracy while using
latent tokens in mechanistically different ways. Probing how latent
tokens are actually used, through marker interventions,
visual-subspace geometry, and attention behavior during latent
generation, is required to distinguish mechanisms that accuracy
alone obscures.

\begin{table}[t]
\centering
\footnotesize
\setlength{\tabcolsep}{1.5pt}
\renewcommand{\arraystretch}{1.15}
\input{tables/tab2_regime_summary}
\caption{
\textbf{Measurements behind the mechanism map.}
Columns group settings by marker-dependence range
(Figure~\ref{fig:mechanism-space}). Marker sensitivity and MOC
track marker reliance; top-5 visual similarity and peak $\Delta H$
track visual-subspace and attention signatures. MOC: fraction of
latent-uniquely-solved examples still solved with markers only.
Full per-setting values in Appendix~\ref{app:concordance}.
}
\label{tab:regime-summary}
\end{table}

%% file: tables/tab2_regime_summary.tex
\begin{tabular}{@{}llll@{}}
\toprule
\textbf{Metric}
& \textbf{Strong}
& \textbf{Reduced}
& \textbf{Weak} \\
\midrule
Settings
& \mvh{}-S/ILVR-S1
& \mvh{}-RL
& ILVR-S2/Monet \\
\midrule
Marker sens.
& 0.59--0.65
& 0.38
& $\leq$0.02 \\
MOC
& 78--100\%
& 91--100\%
& 0--90\% \\
Top-5 sim.
& 0.13--0.55
& 0.10
& 0.07--0.27 \\
Peak $\Delta H$
& $\geq$1.90
& 2.39
& Monet $<$1.0 \\
\bottomrule
\end{tabular}

%% file: 6_conclusion.tex
\section{Conclusion}
\label{sec:conclusion}
Latent visual reasoning is often described as if the slots between
boundary markers carried portable visual evidence. We tested this
account on a favorable probe, \mvh{}, which reaches state-of-the-art
accuracy among the latent-visual settings we compare. Across four
tests, slot contents show little perturbation sensitivity, do not
substantially recover missing visual input or reliably transfer donor
content, and are not required for most of the unique latent-token gain.
The results instead implicate marker and format control, together with attention routing, without exhaustively accounting for the benefit; latent positions may still provide computation, timing, formatting, or routing capacity.
Two diagnostic axes, dependence on boundary markers and proximity
between slot states and visual-token representations, further
separate settings with similar V$^*$ accuracy into mechanistically
distinct uses of the latent tokens, shaped by training supervision.
Latent visual reasoning should therefore be evaluated not only by
accuracy but by what the model actually relies on.

%% file: tables/tabA1_hyperparams.tex
\begin{tabular}{>{\raggedright\arraybackslash}p{0.55\columnwidth}>{\raggedright\arraybackslash}p{0.35\columnwidth}}
\toprule
Quantity & Value / note \\
\midrule
\multicolumn{2}{l}{\emph{Shared}} \\
Backbones & Qwen2.5-VL-7B~\citep{bai2025qwen25vltechnicalreport}, Qwen3-VL-8B~\citep{bai2025qwen3vltechnicalreport} \\
Coverage tokens & $K_g=10$ \\
Informative hypotheses & $K_i=10$ \\
Stage-1 latent budget & $N=K_g+K_i=20$ \\
RL / inference latent budget & $K_{\text{eval}}=8$ \\
Rewrite blend & $\rho=0.6$ \\
EMA teacher decay & $\tau=0.999$ \\
\midrule
\multicolumn{2}{l}{\emph{Stage-1}} \\
Latent-loss target weight & $\lambda_{\text{lat}}^{\star}=1.0$ \\
Latent-loss warm-up & first $10\%$ of steps \\
Text-loss weight & $1.0$ \\
\midrule
\multicolumn{2}{l}{\emph{Stage-2}} \\
GRPO group size & $G=8$ \\
PPO clip range & $\varepsilon=0.2$ \\
PPO epochs per rollout & $E=2$ \\
Reward weights & $\alpha_o=0.9$, $\alpha_f=0.1$ \\
Latent-loss weight & $\lambda_{\text{lat}}^{\text{RL}}=0.5$ \\
Anchor (trust-region) weight & $\beta_{\text{anchor}}=0.1$ \\
\bottomrule
\end{tabular}

%% file: tables/tabA2_full_accuracy.tex
\begin{tabular}{lcccccccccccc}
        \toprule
        \multirow{2}{*}{\textbf{Model}} & \multicolumn{3}{c}{\textbf{V*}} & \multicolumn{3}{c}{\textbf{HRBench4K}} & \multicolumn{3}{c}{\textbf{HRBench8K}} & \multicolumn{3}{c}{\textbf{MME-RealWorld-Lite}} \\
        \cmidrule(lr){2-4}\cmidrule(lr){5-7}\cmidrule(lr){8-10}\cmidrule(lr){11-13}
        & Overall & Attribute & Spatial & Overall & FSP & FCP & Overall & FSP & FCP & Overall & Reasoning & Perception \\
        \midrule
        \rowcolor[HTML]{E1E1E1}
        \multicolumn{13}{c}{\textit{\textbf{Proprietary Model}}}
        \\
        GPT-4o & 67.5* & 72.2* & 60.5* & 59.0* & 70.0* & 48.0* & 55.5* & 62.0* & 49.0* & 52.0* & 48.3* & 54.4* \\
        \midrule
        \rowcolor[HTML]{E1E1E1}
        \multicolumn{13}{c}{\textit{\textbf{Open-Source Model}}}
        \\
        Qwen2.5-VL-7B & 76.4 & 77.4 & 75.0 & 68.0 & 80.3 & 55.8 & 63.8 & 73.8 & 53.8 & 45.8 & 39.7 & 49.6 \\
        \quad + vanilla SFT & 81.7* & 83.5* & 79.0* & 68.4* & 78.3* & \textbf{58.5*} & 61.6* & 70.8* & 52.5* & 51.3* & 46.4* & 54.4* \\
        \quad + vanilla SFT + GRPO & 78.5* & 78.3* & 79.0* & 70.0* & 83.3* & 56.8* & 66.8* & 78.0* & \uline{55.5*} & 52.4* & 48.1* & 55.2* \\
        \rowcolor[HTML]{EEEEEE}
        PixelReasoner & 80.6* & 83.5* & 76.3* & 72.9* & 86.0* & 60.3* & 66.9* & 80.0* & 54.3* & 49.7* & 44.5* & 53.1* \\
        \rowcolor[HTML]{EEEEEE}
        Deepeyes & 83.3* & 84.4* & 81.6* & 71.3* & 83.8* & 58.8* & 65.1* & 77.0* & 53.3* & 54.3* & 50.5* & 56.6* \\
        LVR & 81.7* & \textbf{84.4*} & 77.6* & 70.8* & 83.8* & \uline{57.8*} & 63.8* & 74.5* & 51.5* & 50.6* & 42.7* & 55.7* \\
        ILVR-Stage1 & 80.6 & \uline{84.3} & 75 & 70.0 & 87.5 & 52.5 & 66.5 & 80.3 & 52.8 & 49.2 & 44.0 & 52.5 \\
        ILVR-Stage2 & 80.6 & \uline{84.3} & 75 & 70.3 & 87.8 & 52.8 & 66.6 & \uline{81.8} & 51.5 & 50.4 & 45.3 & 53.6 \\
        Monet-SFT & 79.6 & 81.7 & 76.3 & 70.4 & 84.0 & 57.0 & 63.0 & 78.3 & 49.5 & \uline{52.9} & \uline{48.4} & \uline{55.8}\\
        Monet-RL & \uline{82.2} & 83.5 & 80.3 & 71.0 & 85.8 & 56.3 & 66.0 & 80.0 & 52.0 & 51.2 & 46.5 & 54.2 \\
        \midrule
        \rowcolor{ourblue}
        \multicolumn{13}{c}{\textit{\textbf{Our Model}}}
        \\
        \textbf{MVH-SFT} & \uline{82.2} & 82.6 & \uline{81.6} & \uline{71.5} & \textbf{88.5} & 54.5 & \uline{67.1} & \textbf{82.5} & 51.8 & 50.4 & 45.3 & 53.7 \\
        \textbf{MVH-RL} & \textbf{83.2} & 82.6 & \textbf{84.2} & \textbf{72.4} & \uline{88.3} & 56.5 & \textbf{69.5} & \textbf{82.5} & \textbf{56.5} & \textbf{53.8} & \textbf{49.6} & \textbf{56.5} \\
        \bottomrule
\end{tabular}

%% file: tables/tabA3_comt.tex
\begin{tabular}{l ccccc ccccc}
\toprule
\multirow{2}{*}{\textbf{Methods}}
& \multicolumn{5}{c}{\textbf{COMT}}
& \multicolumn{5}{c}{\textbf{COMT}} \\
\cmidrule(lr){2-6} \cmidrule(lr){7-11}

& Creation & Deletion & Selection & Update & \textbf{Avg.}
& Creation & Deletion & Selection & Update & \textbf{Avg.} \\
\midrule
\textbf{Backbones} & \multicolumn{5}{c}{\textit{Qwen2.5-VL-7B}} & \multicolumn{5}{c}{\textit{Qwen3-VL-8B}} \\
\midrule

\rowcolor[HTML]{E1E1E1}
\multicolumn{11}{c}{\textit{\textbf{Standard Baselines}}} \\

Zero-shot
& 68.0 & 38.0 & 35.0 & 14.0 & 38.8
& \textbf{89.0} & 32.0 & 15.0 & 24.0 & 40.0 \\

Direct-SFT
& 52.0* & 60.0* & \uline{51.0*} & 49.0* & 53.0*
& \textbf{89.0*} & 67.0* & 49.0* & 53.0* & 64.5* \\

CoT-SFT
& \textbf{80.0*} & 52.0* & 45.0* & 46.0* & 55.8*
& 83.0* & 62.0* & 49.0* & 44.0* & 59.5* \\

\midrule

\rowcolor[HTML]{E1E1E1}
\multicolumn{11}{c}{\textit{\textbf{Latent Reasoning}}} \\

Mirage-Stage1
& 53.0* & 54.0* & 45.0* & 42.0* & 48.5*
& 81.0* & 58.0* & 43.0* & 50.0* & 58.0* \\

Mirage-Stage2
& 65.0* & 62.0* & 47.0* & \uline{50.0*} & 56.0*
& 84.0* & 66.0* & 54.0* & 57.0* & 65.3* \\

ILVR-Stage1
& \uline{74.0} & 62.0 & 39.0 & 42.0 & 54.3
& 83.0 & 65.0 & 43.0 & 53.0 & 61.0 \\

ILVR-Stage2
& 68.0 & \textbf{68.0} & 48.0 & 44.0 & 57.0
& 86.0 & 68.0 & 54.0 & 57.0 & 66.3 \\

\midrule
\rowcolor{ourblue}
\multicolumn{11}{c}{\textit{\textbf{Our Model}}} \\

\textbf{MVH-SFT}
& 71.0 & 65.0 & 48.0 & 46.0 & \uline{57.5}
& \uline{88.0} & \uline{70.0} & \uline{59.0} & \uline{61.0} & \uline{69.5} \\

\textbf{MVH-RL}
& 73.0 & \uline{67.0} & \textbf{53.0} & \textbf{53.0} & \textbf{61.5}
& \textbf{89.0} & \textbf{72.0} & \textbf{63.0} & \textbf{64.0} & \textbf{72.0} \\

\bottomrule
\end{tabular}

%% file: tables/tabA4_diagnostic_coverage.tex
\begin{tabular}{@{}>{\raggedright\arraybackslash}p{0.28\textwidth}>{\raggedright\arraybackslash}p{0.34\textwidth}>{\raggedright\arraybackslash}p{0.28\textwidth}@{}}
\toprule
Diagnostic & Settings & Benchmarks \\
\midrule
Core slot/marker perturbation & Original six settings; Mirage-Stage1/2; LVR-SFT & Four multiple-choice benchmarks \\
Marker-only coverage & Original six settings; Mirage-Stage1/2; LVR-SFT & Four multiple-choice benchmarks \\
Cross-example swap & Original six settings & Four multiple-choice benchmarks \\
Image-free injection & Original six settings; bootstrap interval for \mvh-SFT & V$^*$ \\
Visual-subspace and entropy probes & Original six settings & V$^*$ \\
Causal attention intervention & \mvh-SFT, \mvh-RL, ILVR-Stage1 & V$^*$ \\
Open-ended extension & \mvh-SFT, \mvh-RL & TextVQA \\
Inference slot-count sweep & \mvh-SFT, \mvh-RL & Four-benchmark average \\
Matched-data RL control & Monet-RL (original/matched), \mvh-RL & Four multiple-choice benchmarks \\
\bottomrule
\end{tabular}

%% file: tables/tabA5_generality_accuracy.tex
\begin{tabular}{@{}lccccc@{}}
\toprule
Setting & V$^*$ & HR-4K & HR-8K & MME-RW & Avg. \\
\midrule
Mirage-Stage1 & 80.6 & 68.8 & 65.0 & 48.8 & 65.8 \\
Mirage-Stage2 & 81.2 & 70.0 & 65.1 & 49.2 & 66.4 \\
LVR-SFT & 81.5 & 70.5 & 65.4 & 49.8 & 66.8 \\
\bottomrule
\end{tabular}

%% file: tables/tabA6_generality_diagnostics.tex
\begin{tabular}{@{}lccccccc@{}}
\toprule
Setting & Normal & ZeroSlot & RandomSlot & ZeroMarker & Slot-sens. & Marker-sens. & MOC [95\% CI] \\
\midrule
Mirage-Stage1 & 65.8 & 66.0 & 65.4 & 3.2 & +0.1 & +62.6 & 99.1 [97.8, 100.0] \\
Mirage-Stage2 & 66.4 & 66.3 & 65.9 & 66.2 & +0.3 & +0.2 & 86.5 [83.4, 89.3] \\
LVR-SFT & 66.8 & 66.9 & 66.3 & 45.1 & +0.2 & +21.7 & 90.4 [86.9, 93.2] \\
\bottomrule
\end{tabular}

%% file: tables/tabA7_textvqa.tex
\begin{tabular}{@{}lccccccccc@{}}
\toprule
Setting & Normal & ZeroSlot & RandomSlot & $K=0$ & ZeroMarker & Slot-sens. & Marker-sens. & MOC & $n_{\mathrm{gain}}$ \\
\midrule
\mvh-SFT & 81.7 & 81.2 & 82.3 & 80.9 & 17.6 & -0.05 & +64.1 & 99.6 & 3,624 \\
\mvh-RL & 82.4 & 82.9 & 81.7 & 81.6 & 43.5 & +0.10 & +38.9 & 92.7 & 2,280 \\
\bottomrule
\end{tabular}

%% file: tables/tabA8_monet_matched.tex
\begin{tabular}{@{}lccccc@{}}
\toprule
Method & V$^*$ & HR-4K & HR-8K & MME-RW & Avg. \\
\midrule
Monet-RL (original) & 82.2 & 71.0 & 66.0 & 51.2 & 67.6 \\
Monet-RL-matched & 82.2 & 70.8 & 66.6 & 50.2 & 67.5 \\
\mvh-RL & 83.2 & 72.4 & 69.5 & 53.8 & 69.7 \\
\bottomrule
\end{tabular}

%% file: tables/tabA9_slot_sweep.tex
\begin{tabular}{@{}lccccccc@{}}
\toprule
$K_{\mathrm{eval}}$ & 4 & 8 & 10 & 12 & 16 & 20 & Max. $\Delta$ \\
\midrule
\mvh-SFT & 67.1 & 67.8 & 67.4 & 67.2 & 67.2 & 67.7 & 0.7 \\
\mvh-RL & 69.5 & 69.7 & 69.7 & 69.6 & 69.2 & 69.8 & 0.6 \\
\bottomrule
\end{tabular}

%% file: tables/tabA10_ablation_rho.tex
\begin{tabular}{@{}lccccc@{}}
\toprule
\textbf{$\boldsymbol{\rho}$} & \textbf{V$^*$} & \textbf{HR-4K} & \textbf{HR-8K} & \textbf{MME-RW} & \textbf{Avg.} \\
\midrule
0.2 & \textbf{82.4} & \uline{71.3} & \uline{67.1} & \uline{50.2} & \textbf{67.8} \\
0.4 & \textbf{82.4} & \uline{71.3} & 66.8 & \textbf{50.4} & \uline{67.7} \\
\rowcolor{ourblue} \textbf{0.6 (MVH-SFT)} & \uline{82.2} & \textbf{71.5} & \uline{67.1} & \textbf{50.4} & \textbf{67.8} \\
0.8 & 81.0 & 71.2 & \textbf{67.2} & 49.9 & 67.3 \\
1.0 & 80.8 & 70.9 & 66.9 & 50.1 & 67.2 \\
\bottomrule
\end{tabular}

%% file: tables/tabA11_ablation_rl.tex
\begin{tabular}{@{}lccccc@{}}
\toprule
\textbf{Method} & \textbf{V$^*$} & \textbf{HR-4K} & \textbf{HR-8K} & \textbf{MME-RW} & \textbf{Avg.} \\
\midrule
Vanilla GRPO & 80.6 & \textbf{73} & 66.25 & 52.9 & 68.2 \\
\rowcolor{ourblue} \textbf{Ours (\mvh-RL)} & \textbf{83.2} & 72.4 & \textbf{69.5} & \textbf{53.8} & \textbf{69.7} \\
\bottomrule
\end{tabular}

%% file: tables/tabA12_intervention_conditions.tex
\begin{tabular}{>{\raggedright\arraybackslash}p{0.17\linewidth}>{\raggedright\arraybackslash}p{0.12\linewidth}>{\raggedright\arraybackslash}p{0.13\linewidth}>{\raggedright\arraybackslash}p{0.24\linewidth}>{\raggedright\arraybackslash}p{0.21\linewidth}}
\toprule
\textbf{Condition} & \textbf{Markers} & \textbf{Latent slots} & \textbf{Tested factor} & \textbf{Interpretation} \\
\midrule
\textsc{Normal} & kept & kept & Reference latent-span inference & Reference condition \\
$K=0$ & kept & removed & Boundary markers without latent slots & Marker-only contribution \\
\textsc{NoLatent} & disabled & disabled & Clean removal of the latent interface & Latent-format dependence \\
\textsc{ZeroSlot} & kept & zeroed & Slot-state values & Slot-content sensitivity \\
\textsc{RandomSlot} & kept & random & Meaningful slot contents & OOD slot-content perturbation \\
\textsc{FixedSlot} & kept & fixed by position & Input-dependent slot variation & Content-variation control \\
\textsc{ZeroMarker} & zeroed & kept & Boundary-marker input embeddings & Malformed span with latent slots retained \\
\textsc{ZeroAll} & zeroed & zeroed & Joint marker and slot corruption & Malformed span without usable contents \\
\textsc{ForcedSlots} w/o \textsc{Markers} & disabled & forced & Bare latent slots without boundary markers & OOD stress test \\
\bottomrule
\end{tabular}

%% file: tables/tabA13_content_tests.tex
\begin{tabular}{p{0.19\linewidth}p{0.28\linewidth}p{0.25\linewidth}p{0.19\linewidth}}
\toprule
Test & Slot-content prediction & Observed & Verdict \\
\midrule
Slot-content perturbation & Zero/random/fixed slot contents should cause a large drop. & Benchmark-averaged slot-content sensitivity is near zero across \mvh, ILVR, and Monet. & Weak slot-content causality. \\
Latent injection & Injected latents should recover image-conditioned accuracy. & \mvh-SFT recovers 7.8\% of the gap, with a wide CI including zero. & Supporting, not decisive. \\
Image-quality bottleneck & Latents should compensate when visual input is degraded. & With-latent and blind curves show no consistent mid-quality kick-in; the latent contribution is broadly stable across quality levels rather than selectively activated. & No standalone visual buffer. \\
In-distribution swap & Real donor slot states should transfer donor content or answers. & Pooled donor adoption is 5.3--12.7\%; every 95\% CI is below chance. & Little donor content transfer. \\
Marker-only coverage & Slot contents should explain the incremental latent gain. & \kzero{} recovers most Normal-vs-\nolatent{} unique solves. & Gain often covered without slot values. \\
\bottomrule
\end{tabular}

%% file: tables/tabA14_perturbation_robustness.tex
\begin{tabular}{@{}lcc@{}}
\toprule
Setting & Slot-sens. [benchmark range] & Marker-sens. \\
\midrule
\mvh-SFT & +1.3 [$-0.3$, +4.1] & +58.5 \\
\mvh-RL & +1.2 [$-0.3$, +3.6] & +37.8 \\
ILVR-Stage1 & +0.3 [$-0.7$, +1.3] & +65.1 \\
ILVR-Stage2 & +0.6 [$-0.5$, +1.9] & +0.1 \\
Monet-SFT & +2.7 [+1.1, +3.7] & $-0.2$ \\
Monet-RL & +2.4 [+1.6, +3.3] & +1.9 \\
\bottomrule
\end{tabular}

%% file: tables/tabA15_blind_injection.tex
\begin{tabular}{@{}>{\raggedright\arraybackslash}p{0.27\columnwidth}>{\raggedright\arraybackslash}p{0.43\columnwidth}cc@{}}
\toprule
Condition & Image / latent source & $n$ & V$^*$ acc. \\
\midrule
Full-image run & original image + generated latent span & 58 & 0.845 \\
Blind run & blank/zero image; latent disabled & 57 & 0.228 \\
Blind + injected latent & blank/zero image + image-conditioned latent states & 58 & 0.276 \\
\midrule
Injected $-$ Blind & -- & -- & +0.048 \\
Recovery rate & gap fraction & -- & 0.078 \\
\bottomrule
\end{tabular}

%% file: tables/tabA16_latent_swap_sanity.tex
\begin{tabular}{lrrr}
\toprule
Setting & Self-swap@L14 & TrueMk+Slot & $\Delta$ \\
\midrule
\mvh-SFT & 0.842 & 0.842 & +0.000 \\
\mvh-RL & 0.842 & 0.842 & +0.000 \\
ILVR-Stage1 & 0.667 & 0.667 & +0.000 \\
ILVR-Stage2 & 0.825 & 0.842 & -0.018 \\
Monet-SFT & 0.719 & 0.737 & -0.018 \\
Monet-RL & 0.544 & 0.544 & +0.000 \\
\bottomrule
\end{tabular}

%% file: tables/tabA17_pooled_swap.tex
\begin{tabular}{@{}lc@{}}
\toprule
Setting & Pooled donor adoption (\%) [95\% CI] \\
\midrule
\mvh-SFT & 7.7 [4.9, 10.8] \\
\mvh-RL & 9.2 [6.1, 12.7] \\
ILVR-Stage1 & 5.7 [3.1, 8.8] \\
ILVR-Stage2 & 5.3 [2.8, 8.3] \\
Monet-SFT & 11.0 [7.4, 14.9] \\
Monet-RL & 12.7 [8.8, 16.7] \\
\bottomrule
\end{tabular}

%% file: tables/tabA18_marker_covered_gain.tex
\begin{tabular}{lccccc}
\toprule
Setting & V$^*$ & HRBench-4K & HRBench-8K & MME-RW-Lite & Mean \\
\midrule
\mvh-SFT
& 1.00 (18)
& 0.98 (125)
& 0.98 (125)
& 0.78 (308)
& 0.94 \\
\mvh-RL
& 1.00 (7)\textsuperscript{$\dagger$}
& 0.97 (34)
& 0.98 (62)
& 0.91 (58)
& 0.97 \\
ILVR-Stage1
& 1.00 (48)
& 1.00 (162)
& 1.00 (168)
& 1.00 (346)
& 1.00 \\
ILVR-Stage2
& 1.00 (35)
& 0.92 (120)
& 0.91 (133)
& 0.83 (390)
& 0.92 \\
Monet-SFT
& 0.90 (30)
& 0.88 (43)
& 0.87 (61)
& 0.45 (77)
& 0.78 \\
Monet-RL
& 0.50 (2)\textsuperscript{$\dagger$}
& 0.61 (28)
& 0.62 (29)
& 0.00 (2)\textsuperscript{$\dagger$}
& 0.43 \\
\bottomrule
\end{tabular}

%% file: tables/tabA19_attention_intervention.tex
\begin{tabular}{@{}lccccc@{}}
\toprule
Setting & No steering & Transplant & Sharp-random & Oracle & $n$ \\
\midrule
\mvh-SFT & 0 & 47 [33, 62] & 42 [28, 56] & 60 [46, 73] & 45 \\
\mvh-RL & 0 & 39 [17, 61] & 33 [13, 56] & 56 [33, 78] & 18 \\
ILVR-Stage1 & 0 & 36 [26, 47] & 33 [23, 44] & 49 [38, 60] & 78 \\
\bottomrule
\end{tabular}

%% file: tables/tabA20_attention_paired.tex
\begin{tabular}{@{}lcc@{}}
\toprule
Setting & Transplant $-$ sharp-random (pt) & 95\% paired bootstrap CI \\
\midrule
\mvh-SFT & +5 & [$-6$, +16] \\
\mvh-RL & +6 & [$-11$, +23] \\
ILVR-Stage1 & +3 & [$-5$, +11] \\
\bottomrule
\end{tabular}

%% file: tables/tabA21_concordance.tex
\begin{tabular}{lcccccc}
\toprule
 & \multicolumn{2}{c}{Strong marker dep.} & Reduced marker dep. & \multicolumn{3}{c}{Weak marker dep.} \\
\cmidrule(lr){2-3}\cmidrule(lr){4-4}\cmidrule(lr){5-7}
Metric & \mvh-SFT & ILVR-S1 & \mvh-RL & ILVR-S2 & Monet-SFT & Monet-RL \\
\midrule
Marker Sensitivity & 0.585 & 0.651 & 0.378 & 0.001 & -0.002 & 0.019 \\
Malformed Interference & 0.497 & 0.522 & 0.380 & -0.113 & -0.059 & 0.011 \\
Marker-only coverage & 78--100\% & 100\% & 91--100\% & 83--100\% & 45--90\% & 0--62\% \\
Marker identity $\Delta$ & -0.77 & -0.37 & -0.60 & +0.02 & +0.04 & -0.05 \\
Self-loop under marker zero & 100\% & 97\% & 100\% & 0\% & 0\% & 0\% \\
Latent blocks under marker zero & $300\times$ & $14\times$ & $105\times$ & $1\times$ & $1\times$ & $1\times$ \\
Latent--visual top-5 sim, coverage & 0.15 & 0.58 & 0.10 & 0.27 & 0.07 & 0.07 \\
Latent--visual top-5 sim, anchor & 0.11 & 0.52 & 0.10 & 0.24 & 0.07 & 0.06 \\
Latent-step attention entropy & 4.31 & 5.21 & 4.27 & 4.57 & 6.59 & n=2 \\
Peak $\Delta H$ & 2.43 & 1.90 & 2.39 & 2.36 & 0.38 & 0.61 \\
Swap max $|\Delta_{self}|$ & 0.018 & 0.035 & 0.018 & 0.000 & 0.053 & 0.088 \\
Final accuracy, V$^*$/avg & .82/.68 & .81/.67 & .83/.70 & .81/.67 & .80/.67 & .82/.68 \\
\bottomrule
\end{tabular}

%% file: custom.bib
@inproceedings{yang2026machine,
  title={Machine Mental Imagery: Empower Multimodal Reasoning with Latent Visual Tokens},
  author={Yang, Zeyuan and Yu, Xueyang and Chen, Delin and Shen, Maohao and Gan, Chuang},
  booktitle={Proceedings of the IEEE/CVF Conference on Computer Vision and Pattern Recognition},
  year={2026}
}

@misc{dong2026interleavedlatentvisualreasoning,
      title={Interleaved Latent Visual Reasoning with Selective Perceptual Modeling}, 
      author={Shuai Dong and Siyuan Wang and Xingyu Liu and Chenglin Li and Haowen Hou and Zhongyu Wei},
      year={2026},
      eprint={2512.05665},
      archivePrefix={arXiv},
      primaryClass={cs.CL},
      url={https://arxiv.org/abs/2512.05665}, 
}

@inproceedings{wang2026monet,
  title={Monet: Reasoning in Latent Visual Space Beyond Images and Language},
  author={Wang, Qixun and Shi, Yang and Wang, Yifei and Zhang, Yuanxing and Wan, Pengfei and Gai, Kun and Ying, Xianghua and Wang, Yisen},
  booktitle={Proceedings of the IEEE/CVF Conference on Computer Vision and Pattern Recognition},
  year={2026}
}

@inproceedings{
li2026latent,
title={Latent Visual Reasoning},
author={Bangzheng Li and Ximeng Sun and Jiang Liu and Ze Wang and Jialian Wu and Xiaodong Yu and Emad Barsoum and Muhao Chen and Zicheng Liu},
booktitle={The Fourteenth International Conference on Learning Representations},
year={2026},
url={https://openreview.net/forum?id=j84WR5ORsC}
}

@inproceedings{li2026imagination,
  title={Imagination Helps Visual Reasoning, But Not Yet in Latent Space},
  author={Li, You and Chen, Chi and Li, Yanghao and Zeng, Fanhu and Huang, Kaiyu and Xu, Jinan and Sun, Maosong},
  booktitle={Proceedings of the 43rd International Conference on Machine Learning},
  year={2026}
}

@misc{viveiros2026whatsholdinglatentvisual,
      title={What's Holding Back Latent Visual Reasoning?}, 
      author={André G. Viveiros and Nuno Gonçalves and André F. T. Martins and Matthias Lindemann},
      year={2026},
      eprint={2605.18445},
      archivePrefix={arXiv},
      primaryClass={cs.CV},
      url={https://arxiv.org/abs/2605.18445}, 
}

@misc{zhang2026visuallatentsknowsay,
      title={Visual Latents Know More Than They Say: Unsilencing Latent Reasoning in MLLMs}, 
      author={Xin Zhang and Qiqi Tao and Jiawei Du and Moyun Liu and Joey Tianyi Zhou},
      year={2026},
      eprint={2605.02735},
      archivePrefix={arXiv},
      primaryClass={cs.LG},
      url={https://arxiv.org/abs/2605.02735}, 
}

@misc{zhang2025latenttokensthinkcausal,
      title={Do Latent Tokens Think? A Causal and Adversarial Analysis of Chain-of-Continuous-Thought}, 
      author={Yuyi Zhang and Boyu Tang and Tianjie Ju and Sufeng Duan and Gongshen Liu},
      year={2025},
      eprint={2512.21711},
      archivePrefix={arXiv},
      primaryClass={cs.CL},
      url={https://arxiv.org/abs/2512.21711}, 
}

@misc{qin2025chainofvisualthoughtteachingvlmsthink,
      title={Chain-of-Visual-Thought: Teaching VLMs to See and Think Better with Continuous Visual Tokens}, 
      author={Yiming Qin and Bomin Wei and Jiaxin Ge and Konstantinos Kallidromitis and Stephanie Fu and Trevor Darrell and XuDong Wang},
      year={2025},
      eprint={2511.19418},
      archivePrefix={arXiv},
      primaryClass={cs.CV},
      url={https://arxiv.org/abs/2511.19418}, 
}

@inproceedings{chen2026think,
  title={Think with 3D: Geometric Imagination Grounded Spatial Reasoning from Limited Views},
  author={Chen, Zhangquan and Zhang, Manyuan and Yu, Xinlei and Luo, Xufang and Sun, Mingze and Pan, Zihao and An, Xiang and Feng, Yan and Pei, Peng and Cai, Xunliang and Huang, Ruqi},
  booktitle={Proceedings of the IEEE/CVF Conference on Computer Vision and Pattern Recognition},
  year={2026}
}

@misc{wang2026foresttreeslatentsuperposition,
      title={Forest Before Trees: Latent Superposition for Efficient Visual Reasoning}, 
      author={Yubo Wang and Juntian Zhang and Yichen Wu and Yankai Lin and Nils Lukas and Yuhan Liu},
      year={2026},
      eprint={2601.06803},
      archivePrefix={arXiv},
      primaryClass={cs.CL},
      url={https://arxiv.org/abs/2601.06803}, 
}

@misc{ding2026colvrenhancingexploratorylatent,
      title={CoLVR: Enhancing Exploratory Latent Visual Reasoning via Contrastive Optimization}, 
      author={Ziyang Ding and Linjian Meng and Yiming Wu and Yuhan Li and Yuhao Liu and Zhen Zhao},
      year={2026},
      eprint={2605.08802},
      archivePrefix={arXiv},
      primaryClass={cs.CV},
      url={https://arxiv.org/abs/2605.08802}, 
}

@misc{zhang2025latentsketchpadsketchingvisual,
      title={Latent Sketchpad: Sketching Visual Thoughts to Elicit Multimodal Reasoning in MLLMs}, 
      author={Huanyu Zhang and Wenshan Wu and Chengzu Li and Ning Shang and Yan Xia and Yangyu Huang and Yifan Zhang and Li Dong and Zhang Zhang and Liang Wang and Tieniu Tan and Furu Wei},
      year={2025},
      eprint={2510.24514},
      archivePrefix={arXiv},
      primaryClass={cs.CV},
      url={https://arxiv.org/abs/2510.24514}, 
}

@inproceedings{
li2025imagine,
title={Imagine While Reasoning in Space: Multimodal Visualization-of-Thought},
author={Chengzu Li and Wenshan Wu and Huanyu Zhang and Yan Xia and Shaoguang Mao and Li Dong and Ivan Vuli{\'c} and Furu Wei},
booktitle={Forty-second International Conference on Machine Learning},
year={2025},
url={https://openreview.net/forum?id=6vk6Xg24ZC}
}

@inproceedings{
zheng2026deepeyes,
title={DeepEyes: Incentivizing ''Thinking with Images'' via Reinforcement Learning},
author={Ziwei Zheng and Michael Yang and Jack Hong and Chenxiao Zhao and Guohai Xu and Le Yang and Chao Shen and XingYu},
booktitle={The Fourteenth International Conference on Learning Representations},
year={2026},
url={https://openreview.net/forum?id=xUyMXkI958}
}

@misc{hao2025traininglargelanguagemodels,
      title={Training Large Language Models to Reason in a Continuous Latent Space}, 
      author={Shibo Hao and Sainbayar Sukhbaatar and DiJia Su and Xian Li and Zhiting Hu and Jason Weston and Yuandong Tian},
      year={2025},
      eprint={2412.06769},
      archivePrefix={arXiv},
      primaryClass={cs.CL},
      url={https://arxiv.org/abs/2412.06769}, 
}

@inproceedings{goyal2024think,
  title={Think before you speak: Training language models with pause tokens},
  author={Goyal, Sachin and Ji, Ziwei and Rawat, Ankit Singh and Menon, Aditya Krishna and Kumar, Sanjiv and Nagarajan, Vaishnavh},
  booktitle={International Conference on Learning Representations},
  volume={2024},
  pages={27896--27923},
  year={2024}
}

@inproceedings{
pfau2024lets,
title={Let{\textquoteright}s Think Dot by Dot: Hidden computation in transformer language models},
author={Jacob Pfau and William Merrill and Samuel R. Bowman},
booktitle={First Conference on Language Modeling},
year={2024},
url={https://openreview.net/forum?id=NikbrdtYvG}
}

@inproceedings{
darcet2024vision,
title={Vision Transformers Need Registers},
author={Timoth{\'e}e Darcet and Maxime Oquab and Julien Mairal and Piotr Bojanowski},
booktitle={The Twelfth International Conference on Learning Representations},
year={2024},
url={https://openreview.net/forum?id=2dnO3LLiJ1}
}

@inproceedings{NEURIPS2024_fb820110,
 author = {Hu, Yushi and Shi, Weijia and Fu, Xingyu and Roth, Dan and Ostendorf, Mari and Zettlemoyer, Luke and Smith, Noah and Krishna, Ranjay},
 booktitle = {Advances in Neural Information Processing Systems},
 doi = {10.52202/079017-4423},
 editor = {A. Globerson and L. Mackey and D. Belgrave and A. Fan and U. Paquet and J. Tomczak and C. Zhang},
 pages = {139348--139379},
 publisher = {Curran Associates, Inc.},
 title = {Visual Sketchpad: Sketching as a Visual Chain of Thought for Multimodal Language Models},
 url = {https://proceedings.neurips.cc/paper_files/paper/2024/file/fb82011040977c7712409fbdb5456647-Paper-Conference.pdf},
 volume = {37},
 year = {2024}
}

@inproceedings{Man_2025_CVPR,
    author    = {Man, Yunze and Huang, De-An and Liu, Guilin and Sheng, Shiwei and Liu, Shilong and Gui, Liang-Yan and Kautz, Jan and Wang, Yu-Xiong and Yu, Zhiding},
    title     = {Argus: Vision-Centric Reasoning with Grounded Chain-of-Thought},
    booktitle = {Proceedings of the IEEE/CVF Conference on Computer Vision and Pattern Recognition (CVPR)},
    month     = {June},
    year      = {2025},
    pages     = {14268-14280}
}

@InProceedings{Bigverdi_2025_CVPR,
    author    = {Bigverdi, Mahtab and Luo, Zelun and Hsieh, Cheng-Yu and Shen, Ethan and Chen, Dongping and Shapiro, Linda G. and Krishna, Ranjay},
    title     = {Perception Tokens Enhance Visual Reasoning in Multimodal Language Models},
    booktitle = {Proceedings of the IEEE/CVF Conference on Computer Vision and Pattern Recognition (CVPR)},
    month     = {June},
    year      = {2025},
    pages     = {3836-3845}
}

@inproceedings{wu2024v,
  title={{V*}: Guided visual search as a core mechanism in multimodal llms},
  author={Wu, Penghao and Xie, Saining},
  booktitle={Proceedings of the IEEE/CVF Conference on Computer Vision and Pattern Recognition},
  pages={13084--13094},
  year={2024}
}

@inproceedings{wang2025divide,
  title={Divide, conquer and combine: A training-free framework for high-resolution image perception in multimodal large language models},
  author={Wang, Wenbin and Ding, Liang and Zeng, Minyan and Zhou, Xiabin and Shen, Li and Luo, Yong and Yu, Wei and Tao, Dacheng},
  booktitle={Proceedings of the AAAI Conference on Artificial Intelligence},
  volume={39},
  number={8},
  pages={7907--7915},
  year={2025}
}

@inproceedings{
zhang2025mmerealworld,
title={{MME}-RealWorld: Could Your Multimodal {LLM} Challenge High-Resolution Real-World Scenarios that are Difficult for Humans?},
author={YiFan Zhang and Huanyu Zhang and Haochen Tian and Chaoyou Fu and Shuangqing Zhang and Junfei Wu and Feng Li and Kun Wang and Qingsong Wen and Zhang Zhang and Liang Wang and Rong Jin},
booktitle={The Thirteenth International Conference on Learning Representations},
year={2025},
url={https://openreview.net/forum?id=k5VHHgsRbi}
}

@misc{shao2024deepseekmathpushinglimitsmathematical,
      title={DeepSeekMath: Pushing the Limits of Mathematical Reasoning in Open Language Models}, 
      author={Zhihong Shao and Peiyi Wang and Qihao Zhu and Runxin Xu and Junxiao Song and Xiao Bi and Haowei Zhang and Mingchuan Zhang and Y. K. Li and Y. Wu and Daya Guo},
      year={2024},
      eprint={2402.03300},
      archivePrefix={arXiv},
      primaryClass={cs.CL},
      url={https://arxiv.org/abs/2402.03300}, 
}

@inproceedings{NEURIPS2022_9d560961,
 author = {Wei, Jason and Wang, Xuezhi and Schuurmans, Dale and Bosma, Maarten and ichter, brian and Xia, Fei and Chi, Ed and Le, Quoc V and Zhou, Denny},
 booktitle = {Advances in Neural Information Processing Systems},
 editor = {S. Koyejo and S. Mohamed and A. Agarwal and D. Belgrave and K. Cho and A. Oh},
 pages = {24824--24837},
 publisher = {Curran Associates, Inc.},
 title = {Chain-of-Thought Prompting Elicits Reasoning in Large Language Models},
 url = {https://proceedings.neurips.cc/paper_files/paper/2022/file/9d5609613524ecf4f15af0f7b31abca4-Paper-Conference.pdf},
 volume = {35},
 year = {2022}
}

@misc{bai2025qwen25vltechnicalreport,
      title={Qwen2.5-VL Technical Report}, 
      author={Shuai Bai and Keqin Chen and Xuejing Liu and Jialin Wang and Wenbin Ge and Sibo Song and Kai Dang and Peng Wang and Shijie Wang and Jun Tang and Humen Zhong and Yuanzhi Zhu and Mingkun Yang and Zhaohai Li and Jianqiang Wan and Pengfei Wang and Wei Ding and Zheren Fu and Yiheng Xu and Jiabo Ye and Xi Zhang and Tianbao Xie and Zesen Cheng and Hang Zhang and Zhibo Yang and Haiyang Xu and Junyang Lin},
      year={2025},
      eprint={2502.13923},
      archivePrefix={arXiv},
      primaryClass={cs.CV},
      url={https://arxiv.org/abs/2502.13923}, 
}

@misc{bai2025qwen3vltechnicalreport,
      title={Qwen3-VL Technical Report}, 
      author={Shuai Bai and Yuxuan Cai and Ruizhe Chen and Keqin Chen and Xionghui Chen and Zesen Cheng and Lianghao Deng and Wei Ding and Chang Gao and Chunjiang Ge and Wenbin Ge and Zhifang Guo and Qidong Huang and Jie Huang and Fei Huang and Binyuan Hui and Shutong Jiang and Zhaohai Li and Mingsheng Li and Mei Li and Kaixin Li and Zicheng Lin and Junyang Lin and Xuejing Liu and Jiawei Liu and Chenglong Liu and Yang Liu and Dayiheng Liu and Shixuan Liu and Dunjie Lu and Ruilin Luo and Chenxu Lv and Rui Men and Lingchen Meng and Xuancheng Ren and Xingzhang Ren and Sibo Song and Yuchong Sun and Jun Tang and Jianhong Tu and Jianqiang Wan and Peng Wang and Pengfei Wang and Qiuyue Wang and Yuxuan Wang and Tianbao Xie and Yiheng Xu and Haiyang Xu and Jin Xu and Zhibo Yang and Mingkun Yang and Jianxin Yang and An Yang and Bowen Yu and Fei Zhang and Hang Zhang and Xi Zhang and Bo Zheng and Humen Zhong and Jingren Zhou and Fan Zhou and Jing Zhou and Yuanzhi Zhu and Ke Zhu},
      year={2025},
      eprint={2511.21631},
      archivePrefix={arXiv},
      primaryClass={cs.CV},
      url={https://arxiv.org/abs/2511.21631}, 
}

@article{schulman2017proximal,
  title={Proximal policy optimization algorithms},
  author={Schulman, John and Wolski, Filip and Dhariwal, Prafulla and Radford, Alec and Klimov, Oleg},
  journal={arXiv preprint arXiv:1707.06347},
  year={2017}
}

@misc{duan2025vlmevalkitopensourcetoolkitevaluating,
      title={VLMEvalKit: An Open-Source Toolkit for Evaluating Large Multi-Modality Models}, 
      author={Haodong Duan and Xinyu Fang and Junming Yang and Xiangyu Zhao and Yuxuan Qiao and Mo Li and Amit Agarwal and Zhe Chen and Lin Chen and Yuan Liu and Yubo Ma and Hailong Sun and Yifan Zhang and Shiyin Lu and Tack Hwa Wong and Weiyun Wang and Peiheng Zhou and Xiaozhe Li and Chaoyou Fu and Junbo Cui and Jixuan Chen and Enxin Song and Song Mao and Shengyuan Ding and Tianhao Liang and Zicheng Zhang and Xiaoyi Dong and Yuhang Zang and Pan Zhang and Jiaqi Wang and Dahua Lin and Kai Chen},
      year={2025},
      eprint={2407.11691},
      archivePrefix={arXiv},
      primaryClass={cs.CV},
      url={https://arxiv.org/abs/2407.11691}, 
}

@inproceedings{cheng2025comt,
  title={Comt: A novel benchmark for chain of multi-modal thought on large vision-language models},
  author={Cheng, Zihui and Chen, Qiguang and Zhang, Jin and Fei, Hao and Feng, Xiaocheng and Che, Wanxiang and Li, Min and Qin, Libo},
  booktitle={Proceedings of the AAAI Conference on Artificial Intelligence},
  volume={39},
  number={22},
  pages={23678--23686},
  year={2025}
}

@misc{zhu2025surveylatentreasoning,
      title={A Survey on Latent Reasoning}, 
      author={Rui-Jie Zhu and Tianhao Peng and Tianhao Cheng and Xingwei Qu and Jinfa Huang and Dawei Zhu and Hao Wang and Kaiwen Xue and Xuanliang Zhang and Yong Shan and Tianle Cai and Taylor Kergan and Assel Kembay and Andrew Smith and Chenghua Lin and Binh Nguyen and Yuqi Pan and Yuhong Chou and Zefan Cai and Zhenhe Wu and Yongchi Zhao and Tianyu Liu and Jian Yang and Wangchunshu Zhou and Chujie Zheng and Chongxuan Li and Yuyin Zhou and Zhoujun Li and Zhaoxiang Zhang and Jiaheng Liu and Ge Zhang and Wenhao Huang and Jason Eshraghian},
      year={2025},
      eprint={2507.06203},
      archivePrefix={arXiv},
      primaryClass={cs.CL},
      url={https://arxiv.org/abs/2507.06203}, 
}

@inproceedings{NIPS2017_68053af2,
 author = {Tarvainen, Antti and Valpola, Harri},
 booktitle = {Advances in Neural Information Processing Systems},
 editor = {I. Guyon and U. Von Luxburg and S. Bengio and H. Wallach and R. Fergus and S. Vishwanathan and R. Garnett},
 pages = {},
 publisher = {Curran Associates, Inc.},
 title = {Mean teachers are better role models: Weight-averaged consistency targets improve semi-supervised deep learning results},
 url = {https://proceedings.neurips.cc/paper_files/paper/2017/file/68053af2923e00204c3ca7c6a3150cf7-Paper.pdf},
 volume = {30},
 year = {2017}
}

@inproceedings{NEURIPS2025_95c6ae3f,
 author = {Sun, Guohao and Hua, Hang and Wang, Jian and Luo, Jiebo and Dianat, Sohail and RABBANI, MAJID and Rao, Raghuveer and Tao, Zhiqiang},
 booktitle = {Advances in Neural Information Processing Systems},
 editor = {D. Belgrave and C. Zhang and H. Lin and R. Pascanu and P. Koniusz and M. Ghassemi and N. Chen},
 pages = {103739--103762},
 publisher = {Curran Associates, Inc.},
 title = {Latent Chain-of-Thought for Visual Reasoning},
 url = {https://proceedings.neurips.cc/paper_files/paper/2025/file/95c6ae3f3393786203a4b6dcb9df1036-Paper-Conference.pdf},
 volume = {38},
 year = {2025}
}

@inproceedings{NEURIPS2025_0c38f547,
 author = {Su, Alex and Wang, Haozhe and Ren, Weiming and Lin, Fangzhen and Chen, Wenhu},
 booktitle = {Advances in Neural Information Processing Systems},
 editor = {D. Belgrave and C. Zhang and H. Lin and R. Pascanu and P. Koniusz and M. Ghassemi and N. Chen},
 pages = {8222--8251},
 publisher = {Curran Associates, Inc.},
 title = {Pixel Reasoner: Incentivizing Pixel Space Reasoning via Curiosity-Driven Reinforcement Learning},
 url = {https://proceedings.neurips.cc/paper_files/paper/2025/file/0c38f54740062529aa4117a04b583f3c-Paper-Conference.pdf},
 volume = {38},
 year = {2025}
}

@inproceedings{singh2019towards,
  title={Towards vqa models that can read},
  author={Singh, Amanpreet and Natarajan, Vivek and Shah, Meet and Jiang, Yu and Chen, Xinlei and Batra, Dhruv and Parikh, Devi and Rohrbach, Marcus},
  booktitle={2019 IEEE/CVF Conference on Computer Vision and Pattern Recognition (CVPR)},
  pages={8309--8318},
  year={2019},
  organization={IEEE}
}
